\documentclass[default,iicol]{sn-jnl}% Default with double column layout

\usepackage{bbm}
\usepackage{amsmath}
\usepackage{multirow}
\usepackage{amsthm}

\usepackage{graphicx} % DO NOT CHANGE THIS
\usepackage{caption} % DO NOT CHANGE THIS AND DO NOT ADD ANY OPTIONS TO IT
\usepackage{subfigure}
\usepackage{lipsum,booktabs}

\newcommand{\figref}[1]{Fig. \ref{#1}}
\newcommand{\tabref}[1]{Tab. \ref{#1}}
\newcommand{\equref}[1]{Eq. (\ref{#1})}
\newcommand{\secref}[1]{Sec. \ref{#1}}
\newcommand{\algoref}[1]{Alg. \ref{#1}}
\usepackage[T1]{fontenc}

\definecolor{redc}{rgb}{0.858, 0.188, 0.478}
\newcommand{\rc}[1]{\ttfamily\textcolor{redc}{#1}}  % add a "#" before

\definecolor{commentcolor}{RGB}{110,154,155}   % define comment color
\newcommand{\PyComment}[1]{\ttfamily\textcolor{commentcolor}{\# #1}}  % add a "#" before the input text "#1"
\newcommand{\PyCode}[1]{\ttfamily\textcolor{black}{#1}} % \ttfamily is the code font
\usepackage{setspace}

\jyear{2021}%

\theoremstyle{thmstyleone}%
\theoremstyle{thmstyletwo}%

\theoremstyle{thmstylethree}%

\begin{document}

\title[Article Title]{SplitNet: Learnable Clean-Noisy Label Splitting for Learning with Noisy Labels}


\author[1]{\fnm{Daehwan} \sur{Kim}}\email{daehwan85.kim@samsung.com}
\equalcont{These authors contributed equally to this work.}

\author[2]{\fnm{Kwangrok} \sur{Ryoo}}\email{kwangrok21@korea.ac.k}
\equalcont{These authors contributed equally to this work.}

\author[1]{\fnm{Hansang} \sur{Cho}}\email{hansang.cho@samsung.com}

\author[2]{\fnm{Seungryong} \sur{Kim}}\email{seungryong\_kim@korea.ac.kr}

\affil[1]{\orgdiv{Samsung Electro-Mechanics}, \orgname{150}, \orgaddress{\street{Maeyeong-ro}, \city{Yeongtong-gu}, \postcode{Suwon}, \state{Gyeonggi}, \country{Korea}}}

\affil[2]{\orgdiv{Korea University}, \orgname{145}, \orgaddress{\street{Anam-ro}, \city{Seongbuk-gu}, \postcode{Seoul}, \state{Korea}}}
% 150, Maeyeong-ro, Yeongtong-gu, Suwon-si, Gyeonggi-do, Republic of Korea
% 145, Anam-ro, Seongbuk-gu, Seoul, Republic of Korea
% \affil[3]{\orgdiv{Department}, \orgname{Organization}, \orgaddress{\street{Street}, \city{City}, \postcode{610101}, \state{State}, \country{Country}}}


\abstract{Annotating the dataset with high-quality labels is crucial for deep networks' performance, but in real-world scenarios, the labels are often contaminated by noise. To address this, some methods were recently proposed to automatically split clean and noisy labels among training data, and learn a semi-supervised learner in a Learning with Noisy Labels (LNL) framework. However, they leverage a handcrafted module for clean-noisy label splitting, which induces a confirmation bias in the semi-supervised learning phase and limits the performance. In this paper, for the first time, we present a learnable module for clean-noisy label splitting, dubbed SplitNet, and a novel LNL framework which complementarily trains the SplitNet and main network for the LNL task. We also propose to use a dynamic threshold based on split confidence by SplitNet to optimize the semi-supervised learner better. To enhance SplitNet training, we further present a risk hedging method. Our proposed method performs at a state-of-the-art level, especially in high noise ratio settings on various LNL benchmarks. The source code can be found at \href{https://ku-cvlab.github.io/SplitNet/}{https://ku-cvlab.github.io/SplitNet/}}

\keywords{Deep learning, learning with noisy labels, semi-supervised learning, clean-noisy label splitting}



\maketitle

\section{Introduction}\label{sec1}

Deep Neural Networks (DNNs) generally rely on large-scale training data with human-annotated good labels for achieving satisfactory performance~\cite{krizhevsky2012imagenet}. However, due to the high costs and complexity of labeling the data, the labels are often contaminated by noise, and thus many works have strived to develop alternative methods that are robust to label noise, which is often called Learning with Noisy Labels (LNL)~\cite{natarajan2013learning}.  

Recent studies for LNL, in general, have attempted to distinguish clean samples from the noisy dataset using handcrafted methods, e.g., Gaussian Mixture Models (GMMs), and then use these clean samples as labeled samples in the Semi-Supervised Learning (SSL) phase~\cite{li2020dividemix,nishi2021augmentation}. However, the shape of the loss distribution often does not follow the Gaussian distribution~\cite{arazo2019unsupervised}, and data with loss values that are not large or small enough cannot be properly distinguished. Furthermore, the dominant approaches maintain multiple models to avoid the risk attributable to the ability of DNNs to fit arbitrary labels, but this often leads to complicated training procedures~\cite{iscen2022learning}. 
Moreover, in the aforementioned LNL methodology that leverages SSL techniques, the weight of the unlabeled loss, one of the most substantial hyper-parameter, must be adjusted carefully depending on the noise ratio to prevent the model from overfitting. However, the noise ratio is challenging to tease out in a real-world environment, proving to be an unrealistic approach.

\begin{figure}[t]
\centering
\includegraphics[width=\linewidth]{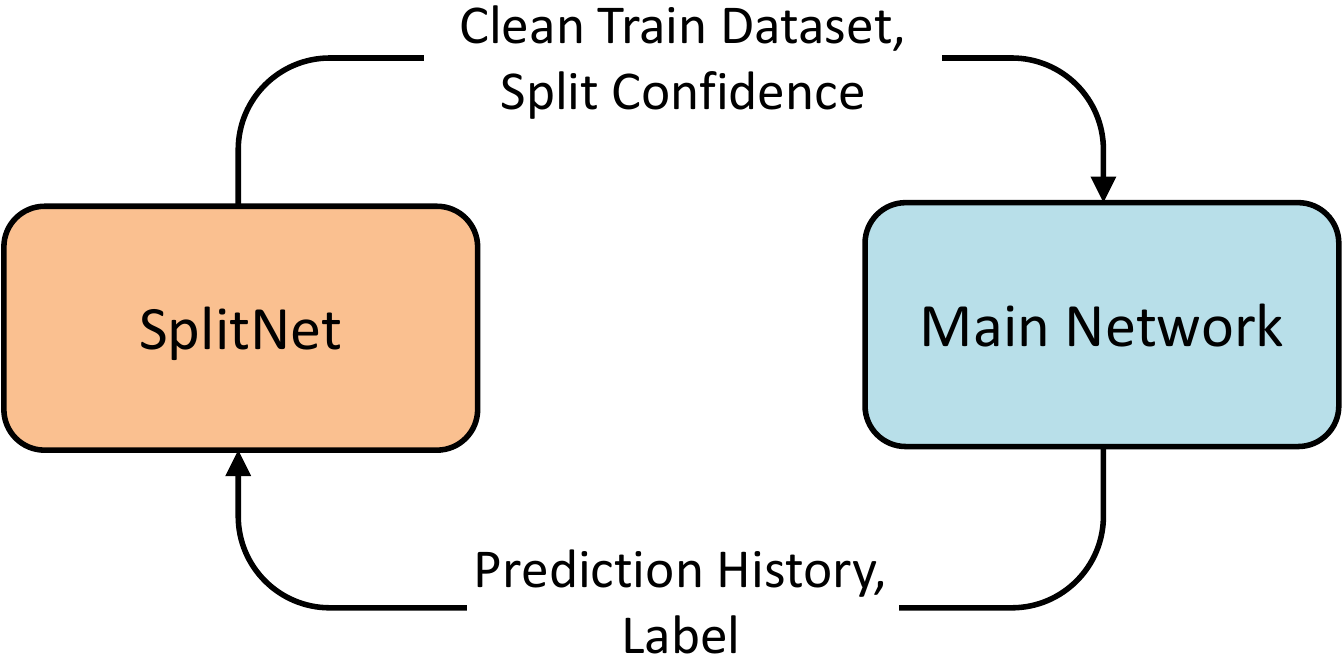}\\
\caption{\textbf{The concept of our alternating update framework with SplitNet.} When the main network outputs the prediction history and label, SplitNet uses them to generate a clean training dataset and deliver it to the main network, which is then used as the labeled data of SSL. These two phases are alternatively used to boost convergence and performance.} 
\label{fig:concept}
\end{figure}

To overcome these limitations, we present a novel framework incorporating a learnable network, called SplitNet, which splits the clean and noisy data in a data-driven manner.
% , which dramatically outperforms conventional methods~\cite{li2020dividemix,nishi2021augmentation} especially in a clean data selection. 
Contrary to conventional methods~\cite{li2020dividemix,nishi2021augmentation} that fit GMMs solely based on per-sample loss distribution to select clean samples, our SplitNet can additionally incorporate the prediction history as input, which allows us to better distinguish ambiguous samples that cannot be precisely distinguished by GMM. In addition, we use a split confidence, a score indicating how confidently SplitNet divides the samples, to determine whether to apply unsupervised loss, enabling more stable learning of SSL method in LNL settings.
% 여기제안된 SSL 이야기가 없음 

More specifically, our overall framework begins with a warm-up and then iteratively learns the main network and SplitNet. As shown in ~\figref{fig:concept}, by formulating the main network and SplitNet in an iterative manner, the two learners are alternately updated, each using the data from the other network. For SplitNet training, the main network provides class prediction and loss distribution, while for the main network training, SplitNet provides split confidences to flexibly adjust the threshold for its SSL procedure. By doing so, whereas previous state-of-the-art methods~\cite{li2020dividemix,nishi2021augmentation} require different hyper-parameter settings for different noise ratios, our proposed model achieves superior performance on all benchmarks, despite its simplicity, requiring only one model and a hyper-parameter setting.

In particular, taking into account the learning status of the main network and the estimated noise ratio of the data set, the thresholds are automatically calculated to distinguish confidently clean and noisy samples. This process which we dub risk hedging, results in a favorable learning environment for SplitNet to mitigate confirmation bias. As the number of confidently clean and noisy sample increase throughout the process, SplitNet enjoys the benefit of a natural curriculum with the aid of the gradually increasing number of hard samples. 

%In order to identify and analyze the factors that led to the success of the proposed model, an extensive ablation study has been included. 

The key contributions of this method are as follows: 

\begin{itemize}
\item{ Our method effectively distinguishes clean samples from noisy datasets compared to other methods through a learnable network called SplitNet.}
% \item{ Our method has achieved excelling performance with a single hyperparameter, regardless of the noise type or ratio, by successfully tailoring the SSL method for LNL settings. }
\item{ As our method enables the learning curriculum to adjust automatically depending on noise ratio, we propose the SSL method that is favorable to LNL by utilizing split confidence obtained through SplitNet. }
\item{ Our method significantly outperforms state-of-the-art results on numerous benchmarks with different types and levels of label noise. %To  validate our method, we also provide extensive ablation study and qualitative results.
}
\end{itemize}

\begin{figure*}[t!]
    \centering
    \includegraphics[width=\linewidth]{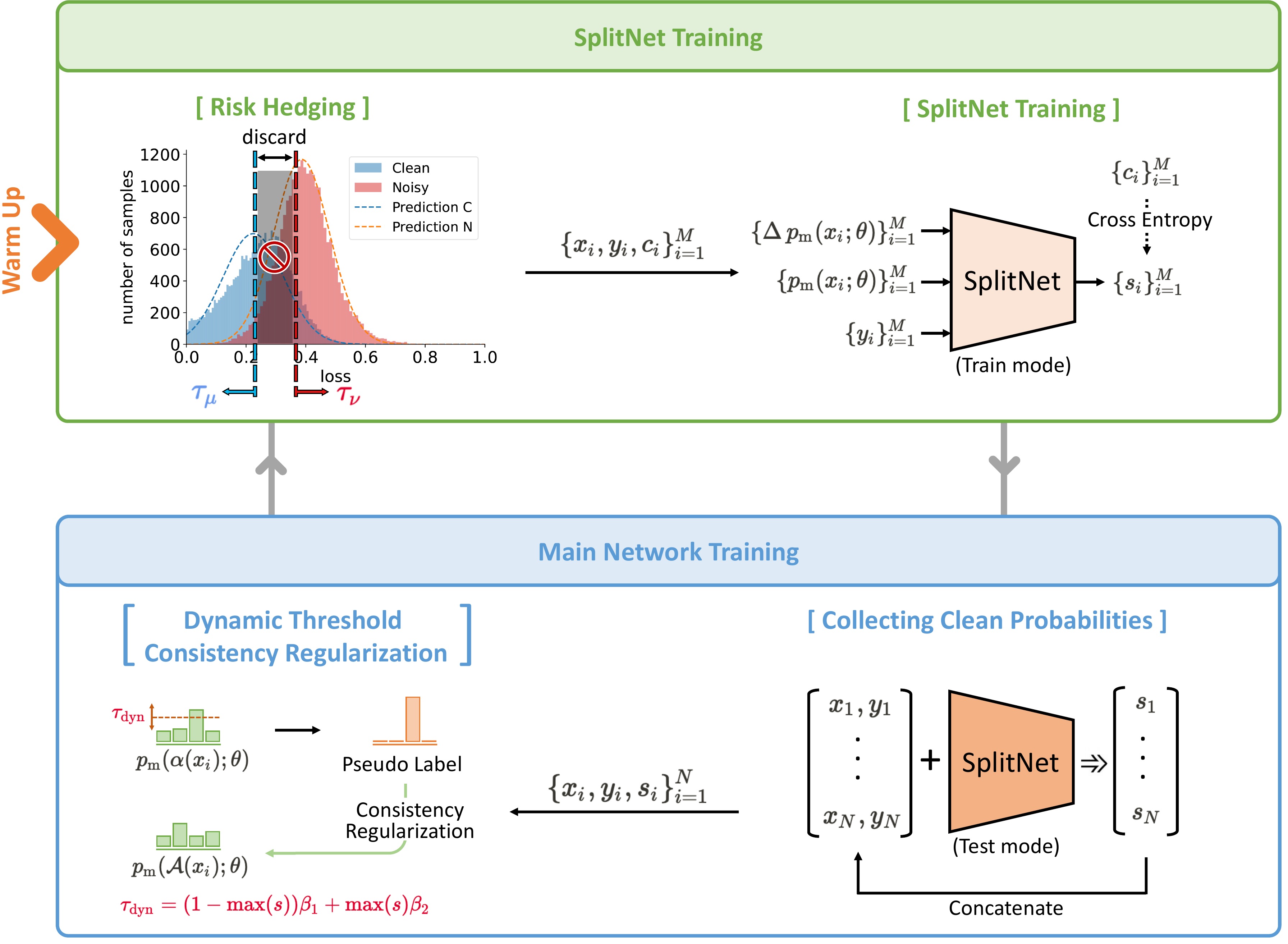}
    \caption{\textbf{The overall architecture of our method.} After training the main model through a warm-up, we use the proposed risk hedging process to only select confident samples to train SplitNet. With SplitNet, we obtain clean probability and split confidence, and with this information, we train the main model through SSL. Loss distribution generated by the main model is used in risk hedging. The main model and SplitNet can be alternately improved through this iterative process.}
    \label{fig:overall} 
    % \vspace{-10pt}
\end{figure*}

\section{Related Work}
\subsection{Learning with Noisy Labels} Modern LNL methods can be largely classified into two categories. The first category uses a loss correction. These methods are further classified into those that relabel noisy samples to correct losses and those that reweights loss depending on each sample. On the one hand, in a study related to the methods that involve relabeling,~\cite{reed2014training} proposed a bootstrap method that adjusts the loss using model prediction. Additionally, the D2L proposed by~\cite{ma2018dimensionality} provided further improvement by using the dimensionality of feature space to determine the weights of the output and label. Furthermore,~\cite{tanaka2018joint} proposed the joint optimization method, which reassigns noisy labels depending on the output of the network, updates networks’ parameters, and labels each epoch. On the other hand, regarding the methods related to reweighting,~\cite{shen2019learning} conducted training by predicting smaller loss samples as clean.

The second category first discards noisy sample labels to apply the semi-supervised learning method. \cite{ding2018semi} and~\cite{kong2019recycling} proved that the SSL method is effective for LNL, and~\cite{li2020dividemix} avoided confirmation bias~\cite{tarvainen2017mean} by having two networks that filter out each other. The~\cite{nishi2021augmentation} studies examined augmentation that was effective for LNL and provided additional related contributions. The method proposed by our paper also utilizes SSL but its novelty compared to existing studies resides in the fact that it only requires a single network to resolve confirmation bias in a data-driven manner using an incidental SplitNet. Moreover, in order to part from a more favorable starting point, the proposed method utilizes $K$-fold cross-filtering to distinguish between clean and noisy data and trains networks using the SSL method.

\subsection{Semi-Supervised Learning} SSL methods aim to utilize not only labeled data but also unlabeled data in order to enhance the performance of a model. SSL methodology is particularly effective when the amount of labeled data is limited and when a large amount of unlabeled data can be used. SSL has been applied in multiple ways in diverse fields of study and is considered a mature research field~\cite{yang2021survey}. In general, SSL methodology can be divided into two areas. These are consistency regularization~\cite{miyato2018virtual,laine2016temporal,tarvainen2017mean}, which forces differently augmented input data to predict the same outcome, entropy minimization~\cite{grandvalet2004semi} and pseudo labeling~\cite{lee2013pseudo}, which allow unlabeled data to produce more confident outcomes. In recent times, a holistic approach that makes use of all of the aforementioned methodologies shows an improved performance~\cite{sohn2020fixmatch,berthelot2019mixmatch,berthelot2019remixmatch}. Most recently, a method that applies a dynamic threshold to consider the varying training circumstances and difficulty levels for different classes~\cite{zhang2021flexmatch,xu2021dash,wang2022freematch} has been identified as the best-performing method.
The method proposed by this paper uses split confidence obtained from SplitNet to dynamize the threshold and to tailor the most avant-garde SSL methods to be used in LNL.

\section{Methodology}
\label{sec:Methods}

\subsection{Overview}

Let us denote $\mathcal{X}= \{(x_i,y_i)\}^N_{i=1}$ as a training dataset, where $x_i$ is an image, $y_i$ is an one-hot label over $r$ classes, and $N$ is the total number of the training data. In the noisy label setting, we assume that $y_i$ could be corrupted, and such labels are called noisy labels. We define noisy data as images with noisy labels. $p_{\mathrm{m}}(x;\theta)$ is the predicted class distribution produced by the main model $p_{\mathrm{m}}(\cdot;\theta)$ with parameters $\theta$ for input $x$. Our goal is to optimize the model parameters $\theta$ so that $p_\mathrm{m}(x_i;\theta)$ approaches the ground-truth label.

\figref{fig:overall} shows our overall architecture. After training the main model through a warm-up, we use the proposed risk hedging process to only select confident samples to train SplitNet. With SplitNet we obtain clean probability and split confidence, and with this information we train the main model through SSL. Loss distribution generated by the main model is used in risk hedging as the whole process is repeated. Through this iterative process, the main model and SplitNet can be alternately improved.

\subsection{SplitNet}
Concretely, given the dataset, the proposed SplitNet is designed to output a probability prediction $s\in\mathbb{R}^2$ regarding the two classes, clean and noisy. The network takes three inputs; model prediction $\{p_{\mathrm{m}}(x_i;\theta)\}_{i}$, the difference in the model predictions of the current and previous iteration $\{\Delta\,p_{\mathrm{m}}(x_i;\theta)\}_{i}$, and one-hot label $\{y_i\}_{i}$ for $i \in \{1,...,M\}$ where $M$ is the total number of samples selected out of a total number of $N$ train data by risk hedging. Note that samples selected by the risk hedging process change in each iteration. In the following, we explain training SplitNet with the proposed risk hedging and semi-supervised learning framework. 

SplitNet is trained to classify the clean and noisy data that has been labeled by GMM; thus, it requires that GMM correctly classifies clean data and noisy data, but there are many cases where the GMM incorrectly classifies data in the overlap between the clean and noisy distributions. A na\"ive solution would be to use a fixed threshold to only select confident data. However, as the model evolves, the loss distribution changes consistently, to which a fixed threshold cannot be adjusted. This leads to the model ignoring a considerable amount of unlabeled data at the earlier stage of training or using a considerable amount of incorrectly labeled data at the late stage of the training~\cite{xu2021dash,zhang2021flexmatch,wang2022freematch}. 

% \vspace{-5pt}
% \paragraph{Training SplitNet with Risk Hedging.} 

% \begin{equation}\label{totalsetforsplitnet_simple}
% \begin{split}
%     &\{(x,y,\mu)|w\ge \tau_{1}\land(x,y,w)\in\mathcal{X}_w\}\\
%     &\cup\{(x,y,\nu)|w\le \tau_{2}\land(x,y,w)\in\mathcal{X}_w\}.
% \end{split}
% \end{equation}

% where $\tau_1$ and $\tau_2$ denotes predefined fixed threshold.

\subsubsection{Risk Hedging} To solve this problem, we propose risk hedging, a process that enhances the training of SplitNet by dynamically adjusting the threshold and selecting confident data. In the risk hedging process, the model’s current learning status and the noise ratio of the training dataset are autonomously determined. A large average value of clean probability distribution implies that the dataset is mostly composed of clean data, and so more overall data can be treated as clean data. A large standard deviation value of clean probability distribution implies that data classification ability is enhanced, and so the next value of the threshold is decreased. Specifically, $\tau_{\mu}$ and $\tau_{\nu}$ should be determined, where $\tau_{\mu}$ denotes the threshold that distinguishes clean data with clean label $\mu$ and  $\tau_{\nu}$ denotes the threshold that distinguishes noisy data with noisy label $\nu$. One-hot label $c\in\{\mu,\nu\}$ is determined by comparing $\tau_{\mu}$ and $\tau_{\nu}$ with the clean probability $w$ derived from the GMM. Formally, for dataset  $\mathcal{X}_w=\{x_i,y_i,w_i\}^N_{i=1}$, the training dataset for SplitNet is defined such that
\begin{equation}\label{totalsetforsplitnet}
\begin{split}
    &\{(x,y,\mu)\mid w\ge \tau_{\mu}\land(x,y,w)\in\mathcal{X}_w\}\\
    &\cup\{(x,y,\nu)\mid w\le \tau_{\nu}\land(x,y,w)\in\mathcal{X}_w\}.
\end{split}
\end{equation}

In the following section, we explain the detailed derivation process of $\tau_{\mu}$ and $\tau_{\nu}$.

\subsubsection{Derivation of \texorpdfstring{$\tau_\mu$}{tau mu} and
\texorpdfstring{$\tau_\nu$}{tau nu}}
\label{sec:derivation}

We define $\mathrm{\tau_{\mu}}$ and $\mathrm{\tau_{\nu}}$ as follows:

\begin{equation}\label{eq:taumunu}
\begin{split}
    \tau_{\mathrm{\mu}} := z-z^\mathrm{F}\!\mu^\mathrm{F}\!\mathrm{P}(\sigma),\\
    \tau_{\mathrm{\nu}} := z-\,z\,\,\mu\,\,\mathrm{P}(\sigma).
    \end{split}
\end{equation}

$\mathrm{P}(\sigma)$ is a function defined as $1-4\,\sigma^2$, and pivot point $z$ is a value between 0 and 1 which serves as a reference point for the clean and noisy thresholds. Each threshold value changes based on $z$. $\mu = \frac{1}{\mid\mathcal{X}\mid}\sum_{i=1}^N{w_i}$ and $\sigma^2 = \frac{1}{\mid\mathcal{X}\mid}\sum_{i=1}^N(w_i-\mu)^2$ are the mean and variance of the clean probability predicted with GMM for the entire dataset, respectively, where $w_i$ is the clean probability of the $i$ th sample predicted with GMM. $\mathrm{F}$ is an operator that performs the following operation where $j$ is an imaginary number:

\begin{equation}\label{zf}
    z^\mathrm{F} := (1-z)j.
\end{equation}

\textbf{lemma 1.} \textit{
Let $x$ be a vector of $n$ numbers in the range [0,$c$], where $c$ is a positive number. Then, the maximum variance of this $n$ number is $c^2/4$.
}
\begin{proof}
Let $\bar{x} = \frac{1}{n}\sum_{i=1}^nx_i$ and $\mathrm{var}(x)=\frac{1}{n}\sum_{i=1}^{n}(x_i-\bar{x})^2$. 
since $x_i \le c$,

$$
\sum\limits_{i}x_i^2=\sum_ix_i\cdot x_i\le\sum_ic\cdot x_i=cn\frac{1}{n}\sum_ix_i =cn\bar{x}. 
$$

Note that $0\le\bar{x}\le c$. Then,

\begin{equation*}
\begin{split}
n\cdot\mathrm{var}(x)&=\sum_i(x_i-\bar{x})^2\\
&=\sum_i(x_i^2-2x_i\bar{x}+\bar{x}^2)\\
&=\sum_ix^2_i-2\bar{x}\sum_ix_i+n\bar{x}^2\\
&=\sum_ix^2_i-2\bar{x}n\frac{1}{n}\sum_ix_i+n\bar{x}^2\\
&=\sum_ix^2_i-n\bar{x}^2\\
&\le cn\bar{x}-n\bar{x}^2=n\bar{x}(c-\bar{x}).
\end{split}
\end{equation*}

And thus

% $$
% \mathrm{var}(x)\le \bar{x}(c-\bar{x}) \le \left(\frac{\bar{x}+(c-\bar{x})}{2}\right)^2=\frac{c^2}{4}.
% $$

$$
\mathrm{var}(x)\le \bar{x}(c-\bar{x}).
$$

Using AM-GM inequality, we get

$$
\bar{x}(c-\bar{x}) \le \left(\frac{\bar{x}+(c-\bar{x})}{2}\right)^2=\frac{c^2}{4}.
$$

This shows that,

$$
\mathrm{var}(x)\le \frac{c^2}{4}.
$$
\end{proof}

\equref{eq:taumunu} is derived as follows. According to lemma 1, for a distribution of real numbers between 0 and 1, the minimum and maximum values of $1-4\sigma^2$ are 0 and 1, respectively. Thus we can set $\tau_\mu$ and $\tau_\nu$ that move dynamically between $z$ and 1, and 0 and $z$, respectively. In this paper, we set the $z$ as 0.5 for all experiments.

\subsection{Network Architecture}
% \subsection{Structure of SplitNet}
\label{sec:structure}

% 우리는 다음과 같은 철학을 따르는 구조를 고안하고자 했다.
% light weight 하면서도 충분한 capacity를 가지도록 하는 모델을 설계하고자 했다.
% 이와 동시에
% 그래서 레이어 층수를 여러개로 만들어보았다.

% splitnet 구조

% \subsubsection{Accuracy According to Structure}
\begin{figure*}[t]
    \centering
    \hspace{-.1in}
    \subfigure[Basic]{
    \includegraphics[width=.32\textwidth]{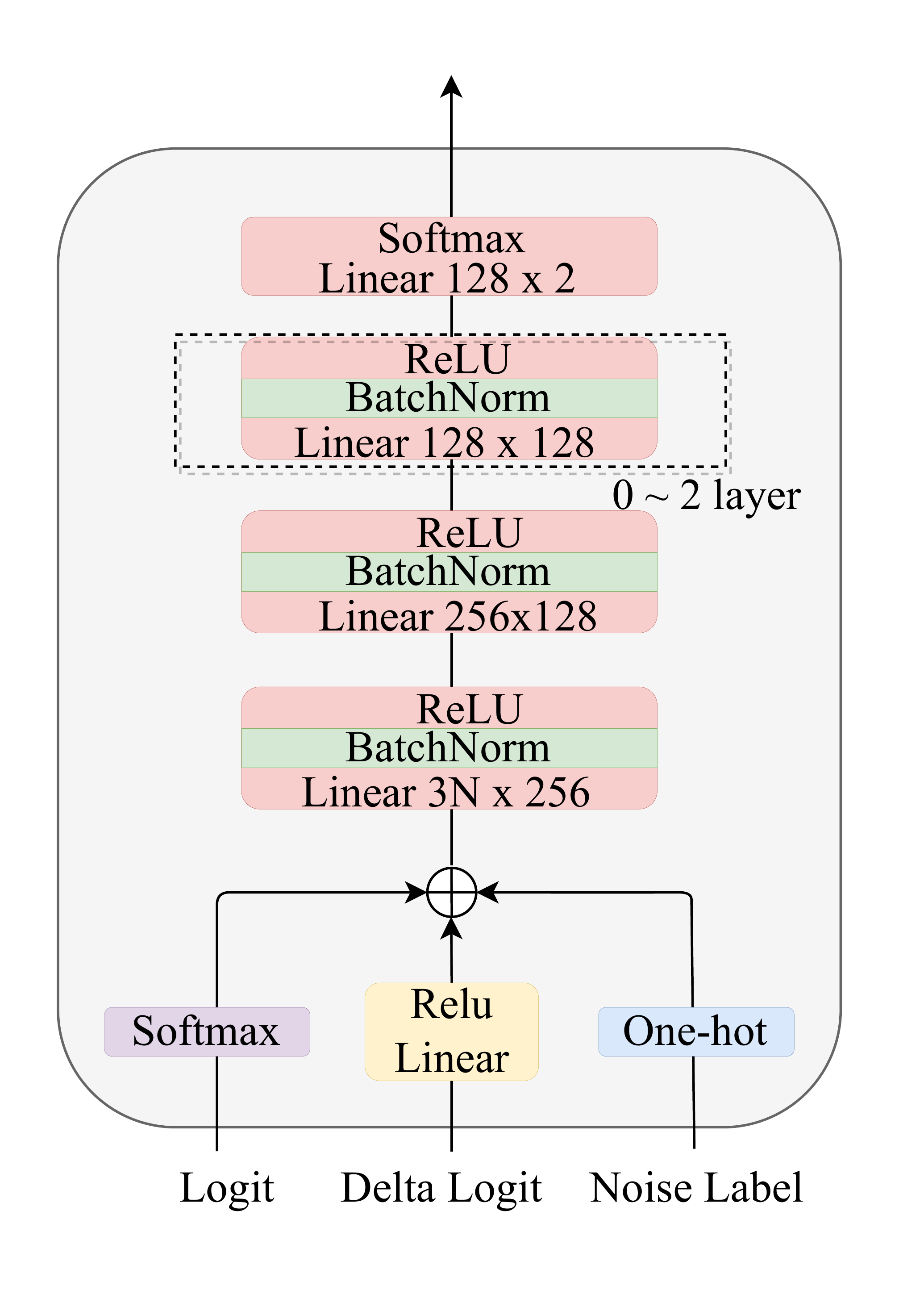}
    \label{fig:org}
    }
    \hspace{-.1in}
    \subfigure[Without prediction difference]{
    \includegraphics[width=.32\textwidth]{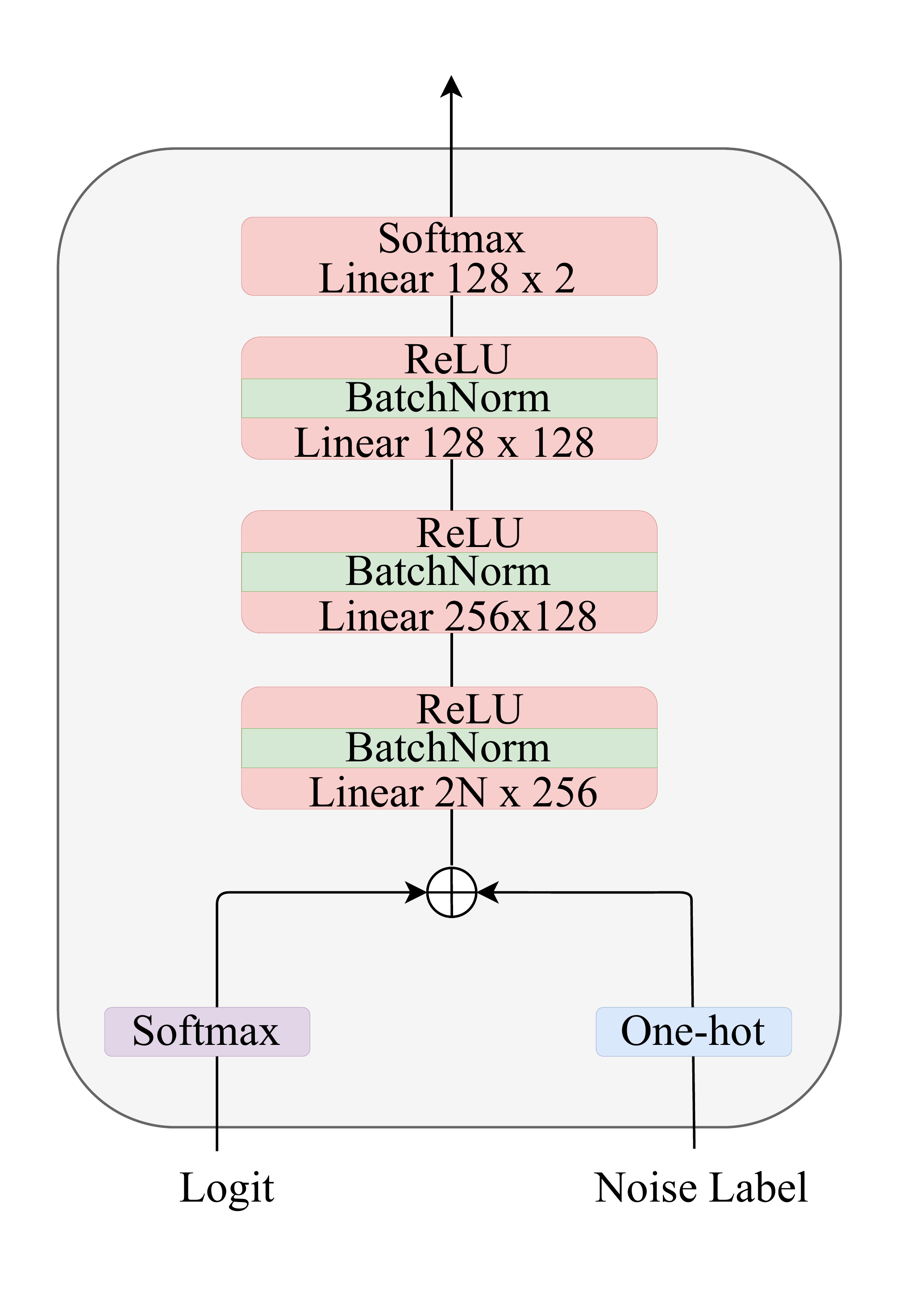}
    \label{fig:nodelta}
    }
    \hspace{-.1in}
    \subfigure[Without batch normalization]{
    \includegraphics[width=.32\textwidth]{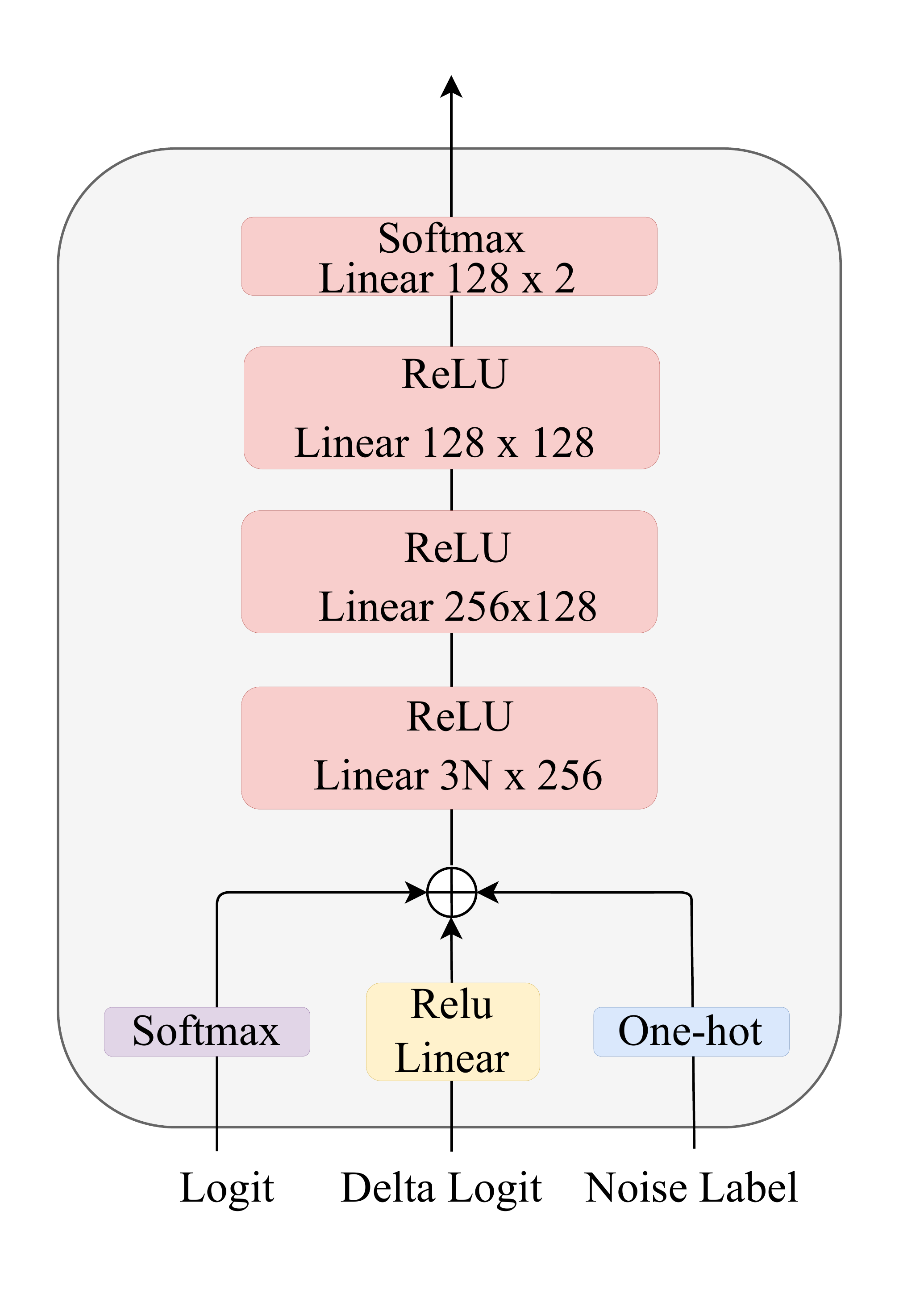}
    \label{fig:nobatchnorm}
    }
    % \vspace{-5pt}
    \caption{\textbf{Variation of SplitNet.} As in (a), the number of layers can be adjusted. (b) is a model that does not consider the prediction difference. (c) is a model without batch normalization.}
    \label{fig:structure}
    % \vspace{-.1in}
\end{figure*}

As shown in \figref{fig:structure}, SplitNet could be implemented in various ways. First of all, we evaluate several modifications of the network architecture to understand SplitNet further. Specifically, we measure the performance of SplitNet by:
\begin{enumerate}
\item Changing the number of layers. (\figref{fig:org})
\item Removing the prediction difference from the input. (\figref{fig:nodelta})
\item Removing the batch normalization~\cite{ioffe2015batch}. (\figref{fig:nobatchnorm})
\end{enumerate}
We experiment with~\figref{fig:org},~\figref{fig:nodelta}, and~\figref{fig:nobatchnorm} for the following reasons.

Samples with noisy labels generate the wrong supervised signal in the warm-up. As discussed in~\secref{sec:warmupstage}, these samples usually have large losses, so during main training, the labels are discarded and learned through unsupervised loss. Therefore, the change in logit per epoch is large, and it can be used as a cue to distinguish noisy data. To confirm this effect, we design a structure~\figref{fig:nodelta} that does not consider logit differences and compare its performance with~\figref{fig:org}.

In addition, in order to design SplitNet to have sufficient capacity while being lightweight, we measure the performance by changing the number of layers as shown in~\figref{fig:org}. The number of layers consisting of~\textsf{Linear - Batch Normalization - ReLU} is increased from 2 to 4.

We evaluate the performance of the SplitNet with batch normalization removed, shown in~\figref{fig:nobatchnorm}, to confirm the importance of batch normalization in the structure of the SplitNet. As a result, convergence fails when batch normalization is not used, verifying the importance of batch normalization.

As a result of the experiment, SplitNet shows the best performance when it is composed of 3 layers based on~\figref{fig:org} with batch normalization, and we adopt this as our structure. We provide a more detailed performance analysis in~\secref{AAS}

\begin{figure}[t]
\centering
% \vspace{-.3in}
% \vspace{-.3in}
\includegraphics[width=\linewidth]{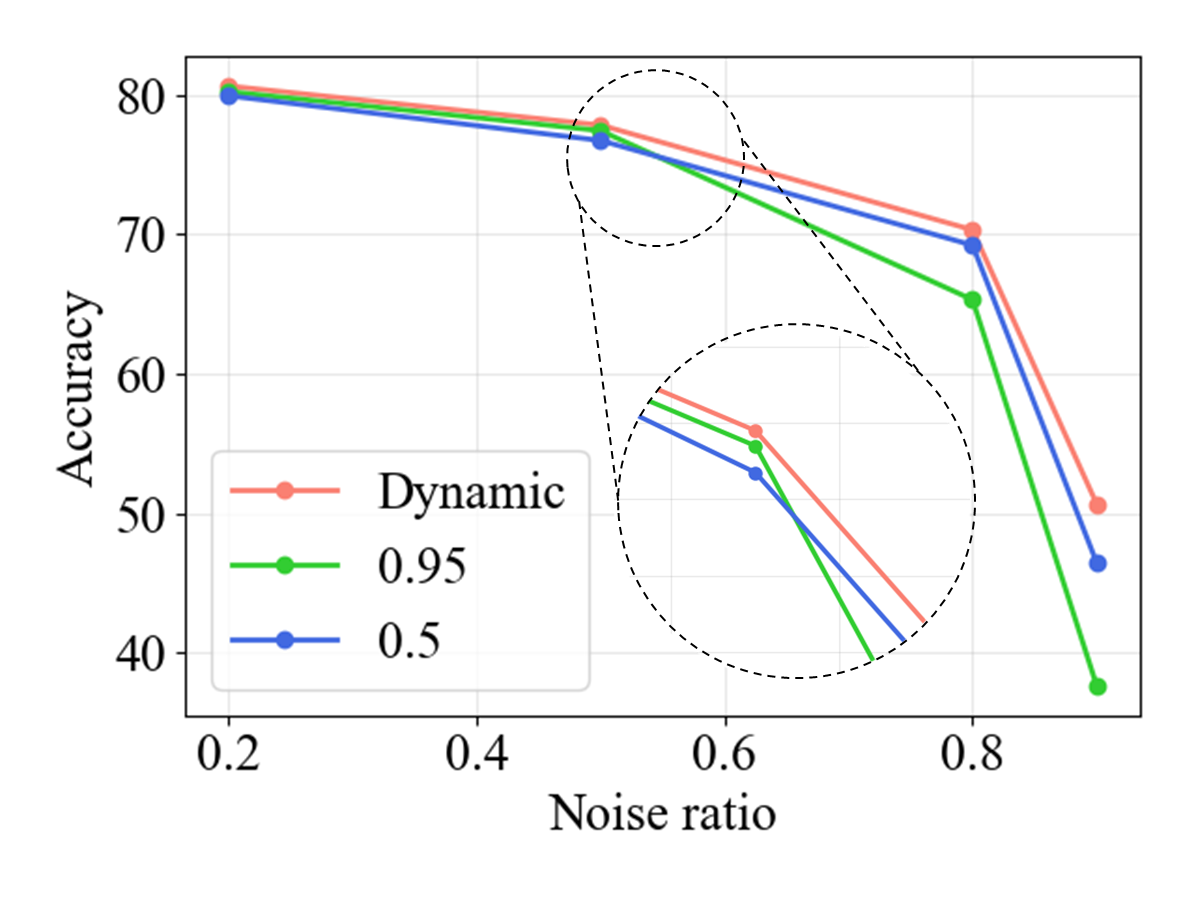}\hfill\\
% \vspace{-5pt}
\caption{\textbf{Pseudo label accuracy by threshold.} The higher the noise ratio, the better the performance at a weak threshold. Conversely, the lower the noise ratio, the better the performance at a strong threshold.}
% \vspace{-10pt}
\label{fig:varythreshold}
\end{figure}

\subsection{Dynamic Thresholding in Semi-Supervised Learner}
To further train the main model, we define the labeled and unlabeled dataset required to train the semi-supervised learner as follows: where $s\in\{s_{\mathrm{clean}},s_{\mathrm{noisy}}\}$ is the binary class prediction with $s_{\mathrm{clean}}$ and $s_{\mathrm{noisy}}$ being the clean and noisy probabilities predicted by SplitNet, dataset $\mathcal{X}$ is forwarded to SplitNet to obtain $s$ and form the dataset $\mathcal{X}_{s}=\{(x_i,y_i,s_i)\}^N_{i=1}$.
% dataset $\mathcal{X}$ is forwarded to SplitNet to obtain $s$ and form the dataset $\mathcal{X}_{s}=\{(x_i,y_i,s_i)\}^N_{i=1}$. Especially, $s\in\{s_{\mathrm{clean}},s_{\mathrm{noisy}}\}$ is the binary class prediction with $s_{\mathrm{clean}}$ and $s_{\mathrm{noisy}}$ being the clean and noisy probabilities predicted by SplitNet. 
Using this dataset, we form a clean labeled dataset  $\mathcal{C}=\{(x,y) \mid s_\mathrm{clean}\ge\tau_{\mathrm{\ label}}\land(x,y,s)\in\mathcal{X}_s\}$ where clean class probability  $s_\mathrm{clean}$ exceeds clean label threshold $\tau_{\ \mathrm{label}}$, and an unlabeled dataset  $\mathcal{U} = \{(x,s) \mid (x,y,s)\in\mathcal{X}_s\}$, which is used for consistency regularization~\cite{rasmus2015semi,sajjadi2016regularization} based learning.
    
Based on these datasets, the semi-supervised loss function consists of two cross-entropy loss terms: supervised loss $\mathcal{L_C}$ and unsupervised loss $\mathcal{L_U}$. First of all, $\mathcal{L_C}$ is the standard cross-entropy loss $\mathcal{H}(\cdot)$ on dataset $\mathcal{C}$ as follows:
\begin{equation}\label{lc}
    \mathcal{L}_\mathcal{C} = \frac{1}{\mathcal{\mid C \mid}}\sum_{(x,y) \in \mathcal{C}}\mathcal{H}(y,p_{\mathrm{m}}(x;\theta)).
\end{equation}
For the unsupervised loss function, we exploit consistency regularization loss, a function used by FixMatch~\cite{sohn2020fixmatch}, one of the most prevalent modern SSL frameworks. However, Our methodology is different in that it maximizes the effect according to the LNL by flexibly adjusting the threshold for determining a stable sample using split confidence that indicates the degree of distance from the decision boundary that divides the clean and noisy samples obtained through SplitNet.
% However, our methodology differs in that we dynamically control the threshold that determines stable samples with clean probability, maximizing effectiveness with LNL. 

% \begin{figure*}[t!]
%     \centering
%     \includegraphics[width=.45\linewidth]{figs/Thresold_acc/Thresold_acc_1.png}
%     \caption{\textbf{Vary in accuracy by threshold.} The higher the noise ratio, the better the performance at a weak threshold. Conversely, the lower the noise ratio, the better the performance at the strong threshold.}
%     \label{fig:thresholdeffect} \vspace{-10pt}
% \end{figure*}

As shown in ~\figref{fig:varythreshold}, in the case of a fixed threshold, in situations with a very high level of label noise, a lower threshold achieves better performance and vice versa. This tendency is the reason that achieving superior performance in all noise ratio benchmarks with only one hyper-parameter setting is a difficult task. With motivation from these findings, we propose a dynamic threshold that is adjusted according to the split confidence of the sample. Our dynamic threshold consistently shows higher accuracy on any noise ratio.

Specifically, we first generate an artificial pseudo-label $q=\mathcal{E}(\mathrm{arg\,max}(p_{\mathrm{m}}(\alpha(x);\theta)))$, where $\alpha(\cdot)$ is a weak augmentation function that can carry out simple transformations (for example, flip and shift) on an image, and $\mathcal{E}$ is a function that one-hot-encodes an index value. Then we enforce the model so that the model output of strongly-augmented data and of weakly-augmented data are consistent.
\begin{equation}\label{lu}
\begin{split}
    \mathcal{L}_{\mathcal{U}}=\frac{1}{\mid \mathcal{U}\mid} \sum_{(x,s)\in\mathcal{U}}&\mathbbm{1}(\mathrm{max}(p_{\mathrm{m}}(\alpha(x);\theta))\\
    &\ge\tau_{\mathrm{dyn}})\mathcal{H}(q,p_{\mathrm{m}}(\mathcal{A}(x);\theta)),
\end{split}
\end{equation}
where $\tau_{\mathrm{dyn}}$ is the dynamically-changing threshold depending on the sample’s split confidence:
\begin{equation}\label{taudynamic}
    \tau_{\mathrm{dyn}} =(1-\mathrm{max}(s))\beta_{1} +\mathrm{max}(s)\beta_{2},
\end{equation}
where $\beta_1$ and $\beta_2$ refer to the upper bound and lower bound of $\tau_{\mathrm{dyn}}$ respectively, and $\mathcal{A}(\cdot)$ the strong augmentation function, which carries out more complex transformations (e.g., RandAug~\cite{cubuk2020randaugment}) on an image. In this way, even without adjustments in hyper-parameters, robust performance is achieved in various noise ratios of the training dataset.

The semi-supervised loss used to train the model can be written as: \begin{equation}\label{totall}
    \mathcal{L}={\eta}\mathcal{L}_\mathcal{C}+(1-\eta)\mathcal{L}_\mathcal{U},
\end{equation}
where $\eta = \mid\mathcal{\,C\,}\mid / \mid\mathcal{\,X\,}\mid$ is a weight automatically adjusted to become smaller as the estimated noise ratio of the dataset is smaller. As a result, the more noisy the dataset, the more unsupervised loss contributes to the total loss.

As shown in~\algoref{algo:pseudo}, we outline our main training algorithm in PyTorch~\cite{paszke2019pytorch} style. In the algorithm, $\theta_s$ denotes the parameters of SplitNet.

\begin{algorithm}[t]
\SetAlgoLined
\small
\begin{spacing}{1.25}
    % \PyComment{this is a comment} \\
    % \PyComment{this is a comment} \\
    % \PyComment{} \\
    % \PyComment{going to have indentation} \\
    % \PyCode{for i in range(N):} \\
    % \Indp   % start indent
    %     \PyComment{your comment} \\
    %     \PyCode{your code} \PyComment{inline comment} \\ 
    % \Indm % end indent, must end with this, else all the below text will be indented
    % \PyComment{this is a comment} \\
    % \PyCode{your code}
\Indp    
    \PyComment{obtain clean probability by GMM} \\ \PyCode{1: $\mathcal{W}$ = GMM($\mathcal{X}$,$\theta$)}  \\
    \PyComment{train SplitNet with confident samples} \\ \PyCode{2: $\theta_s$ = AdamW(RiskHedging($\mathcal{X}$,$\mathcal{W}$),$\theta_s$)} \\
    \PyComment{obtain clean probability by SplitNet} \\ \PyCode{3: $\mathcal{S}$ = SplitNet($\mathcal{X}$,$\theta_s$)}  \\
    \PyComment{generate pseudo labels} \\ \PyCode{4: $q$ = one-hot$($arg\,max$(p_\mathrm{m}(\alpha(x);\theta)))$}   \\
    \PyComment{set dynamic thresholds} \\ \PyCode{5: $\tau_{\mathrm{dyn}}$ = (1-max($\mathcal{S}$))$\beta_1$+max($\mathcal{S}$)$\beta_2$}  \\
    \PyCode{6: $\mathcal{L_U}$ = $(1/\mid\mathcal{X}\mid) \sum_\mathcal{X} \mathbbm{1}$(max($p_\mathrm{m}(\alpha(x);\theta))\ge\tau_{\mathrm{dyn}}$)\\$\mathrm{H}(q,p_\mathrm{m}(\mathcal{A}(x);\theta))$} \\
    \PyCode{7: $\mathcal{C}$ = $\{(x,y)\mid s_{\mathrm{clean}}\ge\tau_{\mathrm{label}},(x,y,s)\in (\mathcal{X},\mathcal{S})\}$} \\
    \PyCode{8: $\mathcal{L_C}$ = $(1/\mid\mathcal{C}\mid)\sum_\mathcal{C}\mathrm{H}(y,p_\mathrm{m}(x;\theta))$} \\
    \PyCode{9: $\eta = \mid\mathcal{C}\mid/\mid\mathcal{X}\mid$} \\
    \PyCode{10: \!\!\!\rc{return} $\eta \mathcal{L_C} + (1-\eta)\mathcal{L_U}$} \\
\Indm
\end{spacing}
\caption{Network Training with SplitNet}
% \caption{SplitNet Training}
\label{algo:pseudo}
\end{algorithm}

\begin{figure*}[t!]
    \centering
    \includegraphics[width=\linewidth]{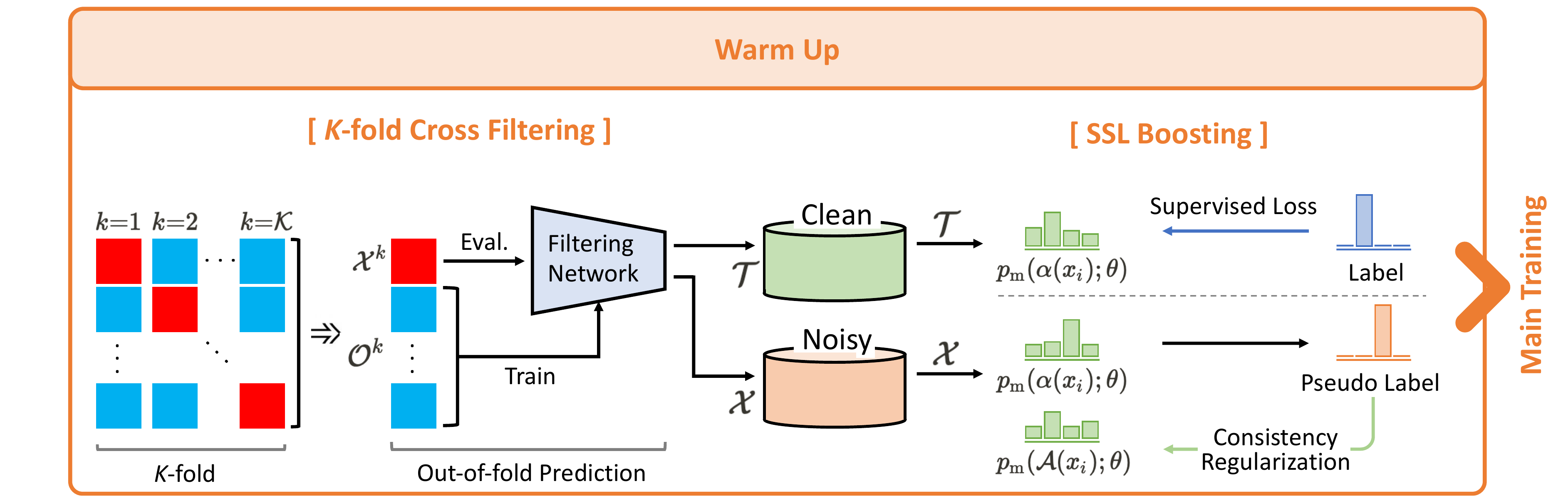}
    \caption{\textbf{The warm-up process.} The SSL learner warms up the model, using clean data selected by $K$-Fold cross-filtering as labeled data.}
    \label{fig:warmup} 
%   \vspace{-10pt}
\end{figure*}

\subsection{Warm-Up Stage}
\label{sec:warmupstage}

%%%%%%%%%%%%%%%%%%%%%%%%%%%%%%%%%%%%%%%%%%%%%%%%%%%%%%%%%%%%%%

In DNN, correctly labeled data tend to converge more quickly than incorrectly labeled data~\cite{arpit2017closer}, which allows samples with lower loss and higher loss to be categorized as clean data and noisy data, respectively. 
In the previous state-of-the-art methods~\cite{li2020dividemix,nishi2021augmentation}, for the initial convergence of the algorithm, the model is trained for a few epochs on a training dataset by using the standard cross-entropy loss. However, this training method does not function effectively in asymmetric noise settings and thus requires the addition of negative entropy loss terms and so forth~\cite{li2020dividemix,nishi2021augmentation,chen2021two}. Performance is also unstable in settings with a high noise ratio. To address this issue, we propose a novel warm-up method that does not require hyper-parameter changes or negative entropy loss because it works well even in high ratio noisy settings or asymmetric noise.~\figref{fig:warmup} shows the diagram of our warm-up.
% Our diagram of warm up is shown in~\figref{fig:warmup}
% \vspace{-5pt}

\begin{figure*}[th!]
    \centering
    \hspace{-.1in}
    % \vspace{-5pt}
    \subfigure[Cross-entropy at 0.2]{
    \includegraphics[width=.24\linewidth]{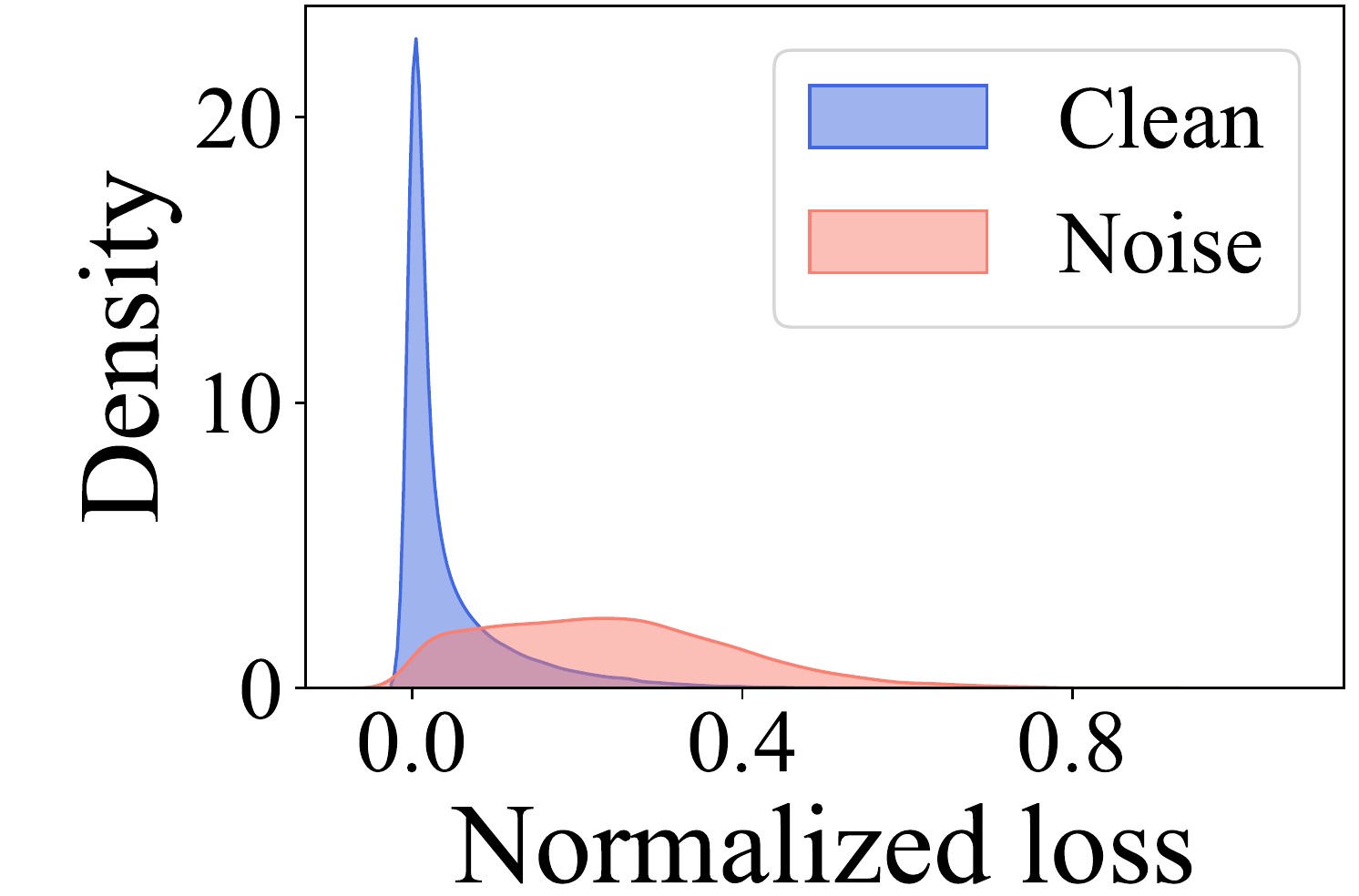}
    \label{fig:20cross}
    }
    \hspace{-.1in}
    \subfigure[Proposed warm-up at 0.2]{
    \includegraphics[width=.24\linewidth]{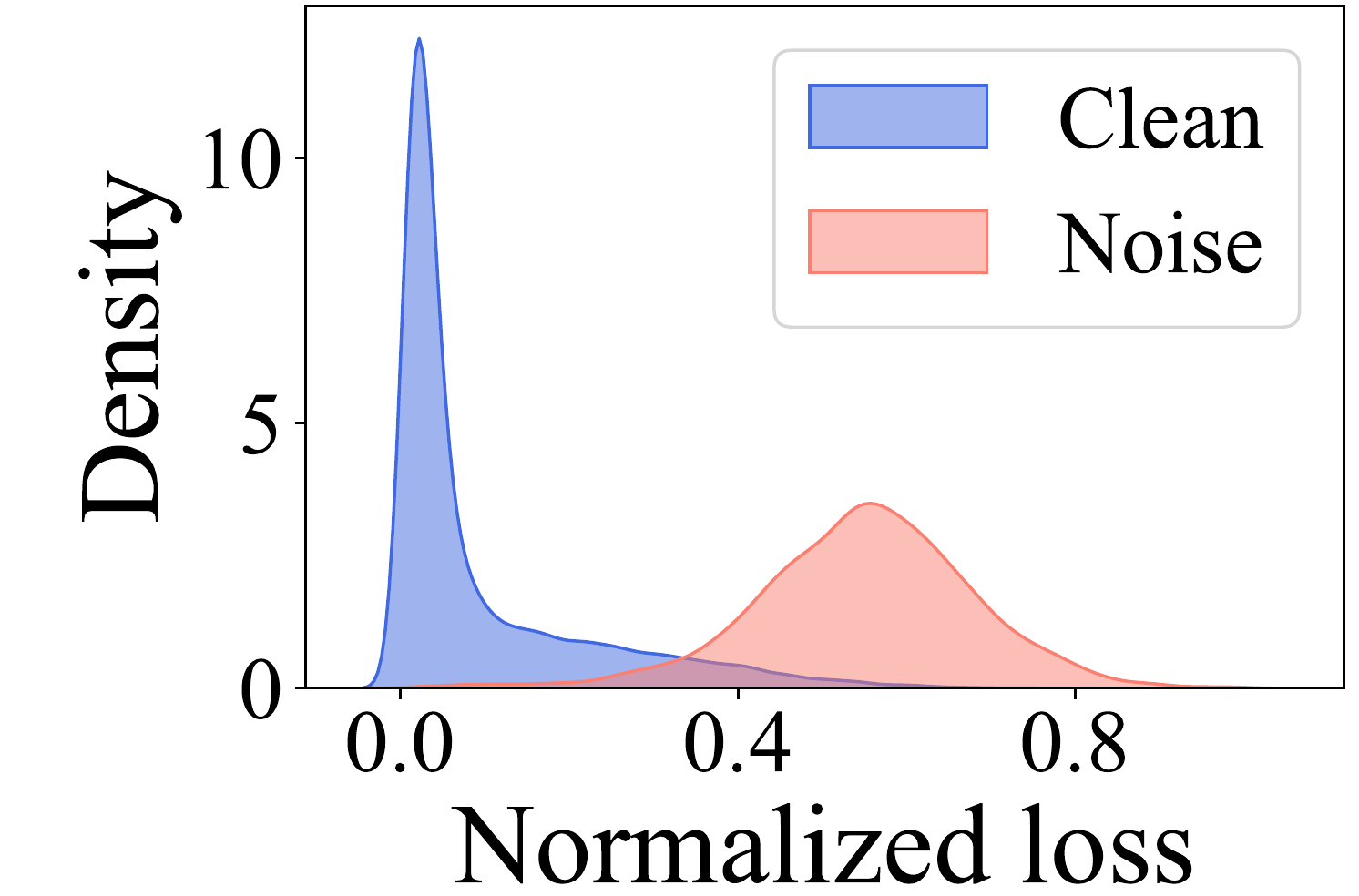}
    \label{fig-20split}
    }
    \hspace{-.1in}
    \subfigure[Cross-entropy at 0.5]{
    \includegraphics[width=.24\linewidth]{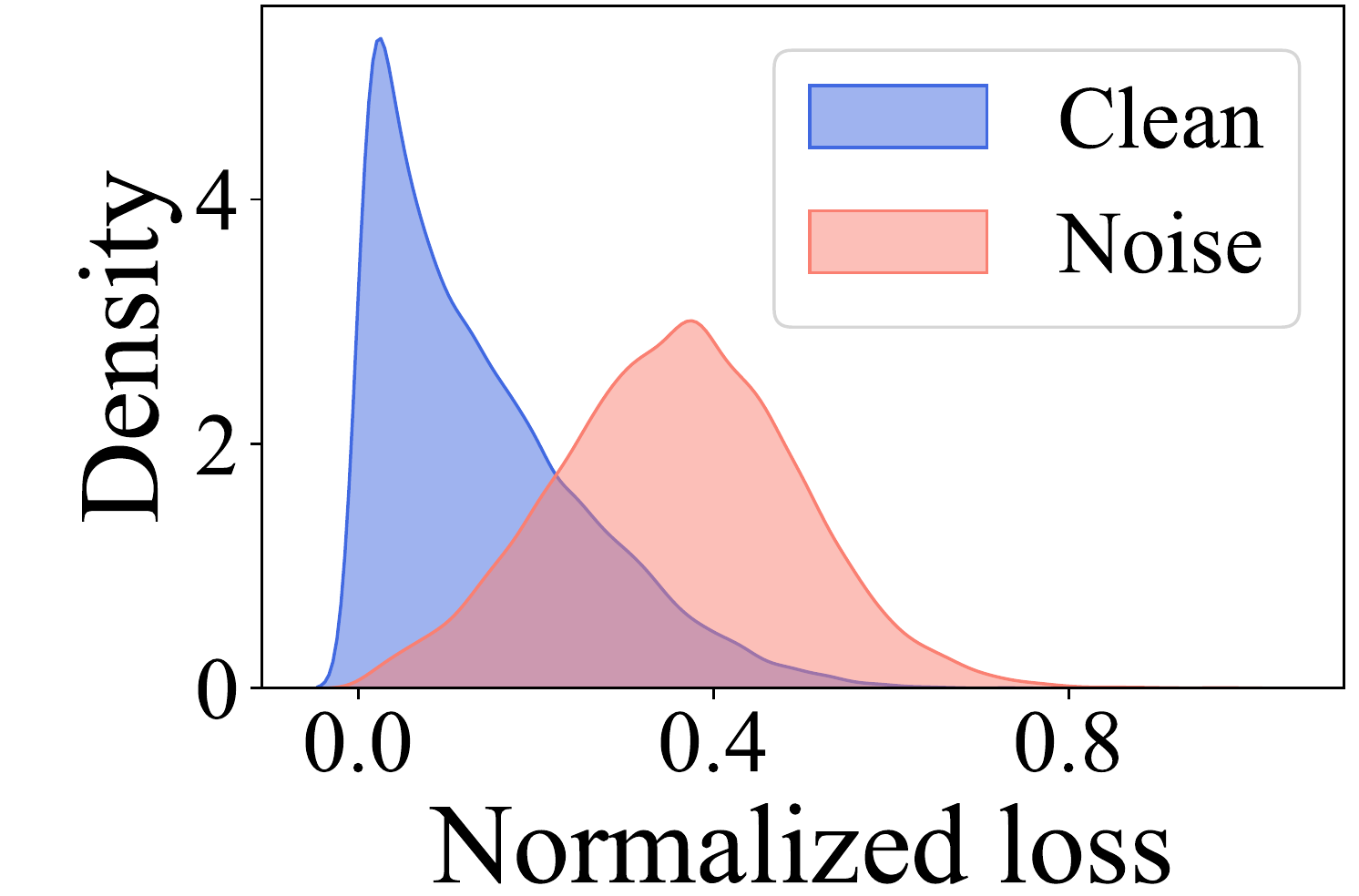}
    \label{fig-50cross}
    }
    \hspace{-.1in}
    \subfigure[Proposed warm-up at 0.5]{
    \includegraphics[width=.24\linewidth]{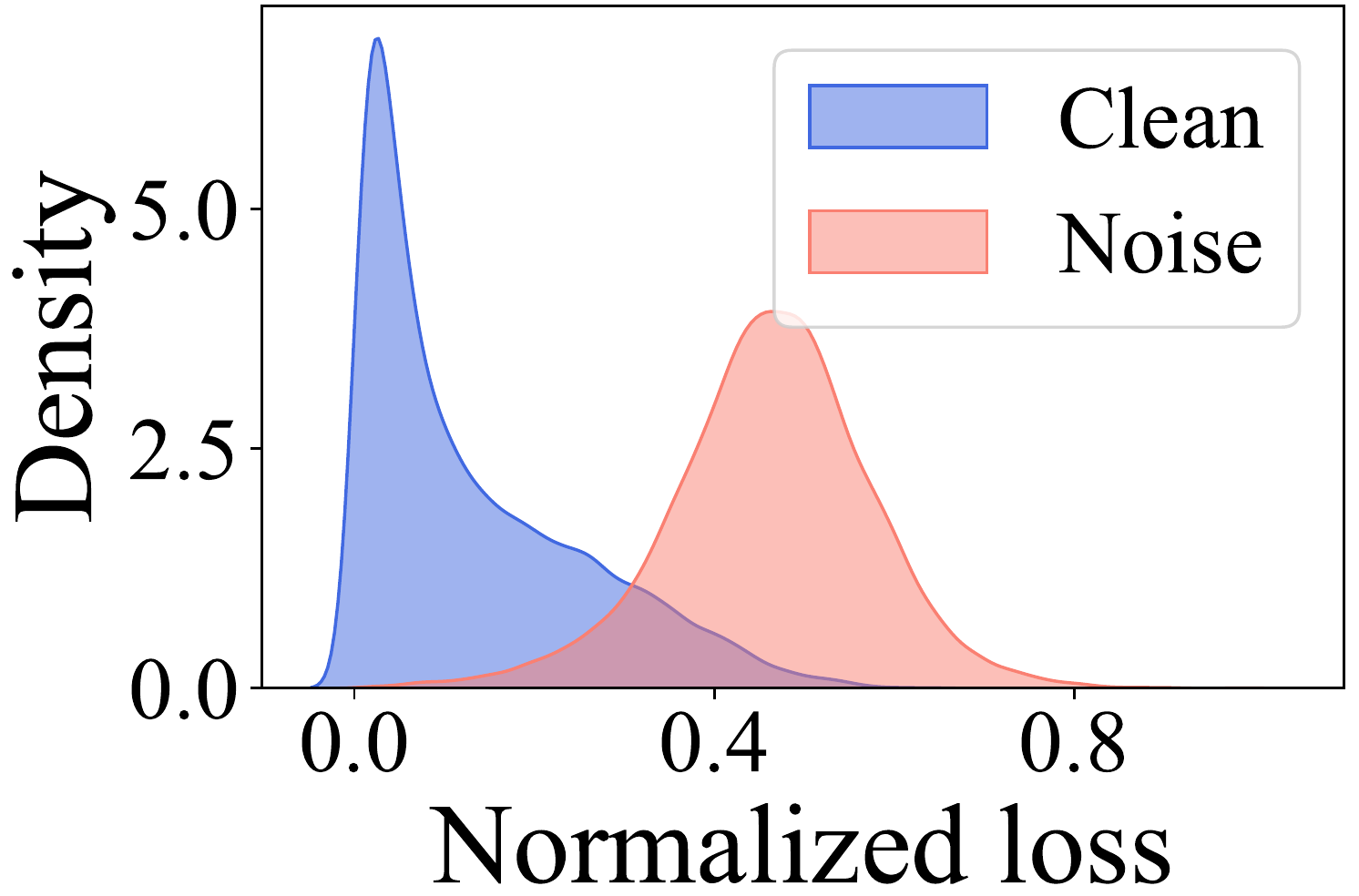}
    \label{fig-50split}
    }\\
    \hspace{-.1in}
    \subfigure[Cross-entropy at 0.8]{
    \includegraphics[width=.24\linewidth]{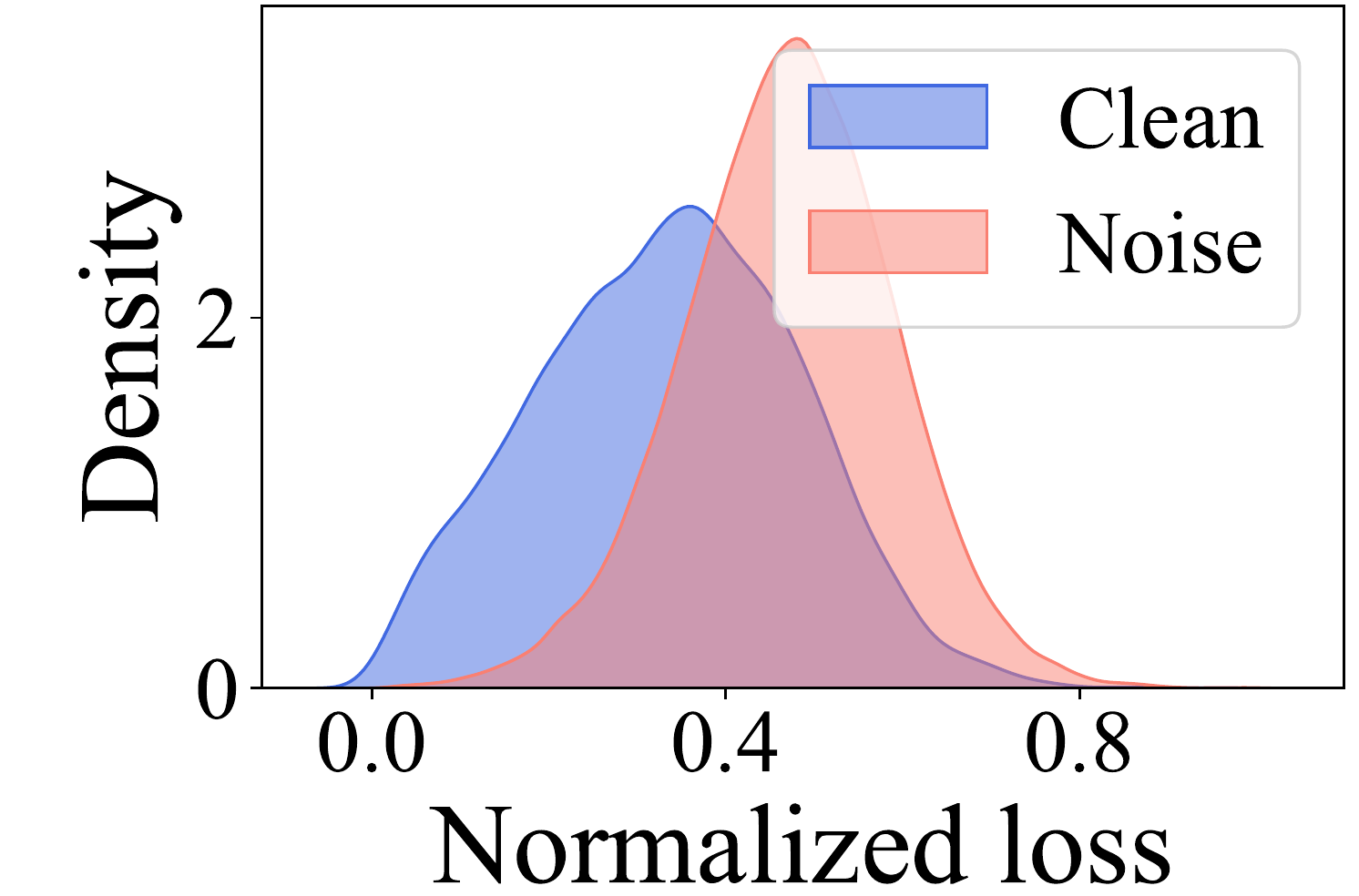}
    \label{fig-80cross}
    }
    \hspace{-.1in}
    \subfigure[Proposed warm-up at 0.8]{
    \includegraphics[width=.24\linewidth]{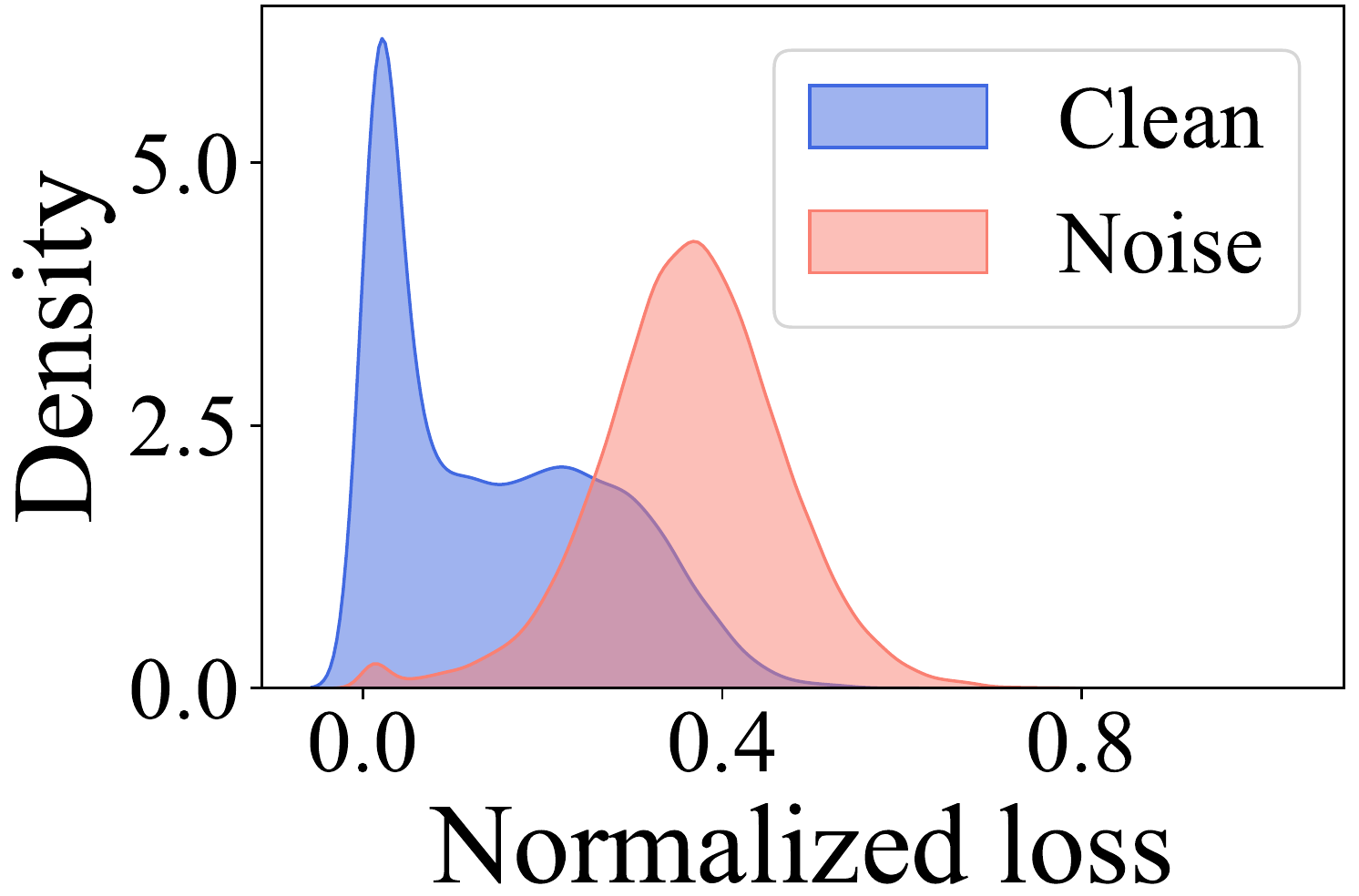}
    \label{fig-80split}
    }
    \hspace{-.1in}
    \subfigure[Cross-entropy at 0.9]
    % \vspace{-5pt}
    {
    \includegraphics[width=.24\linewidth]{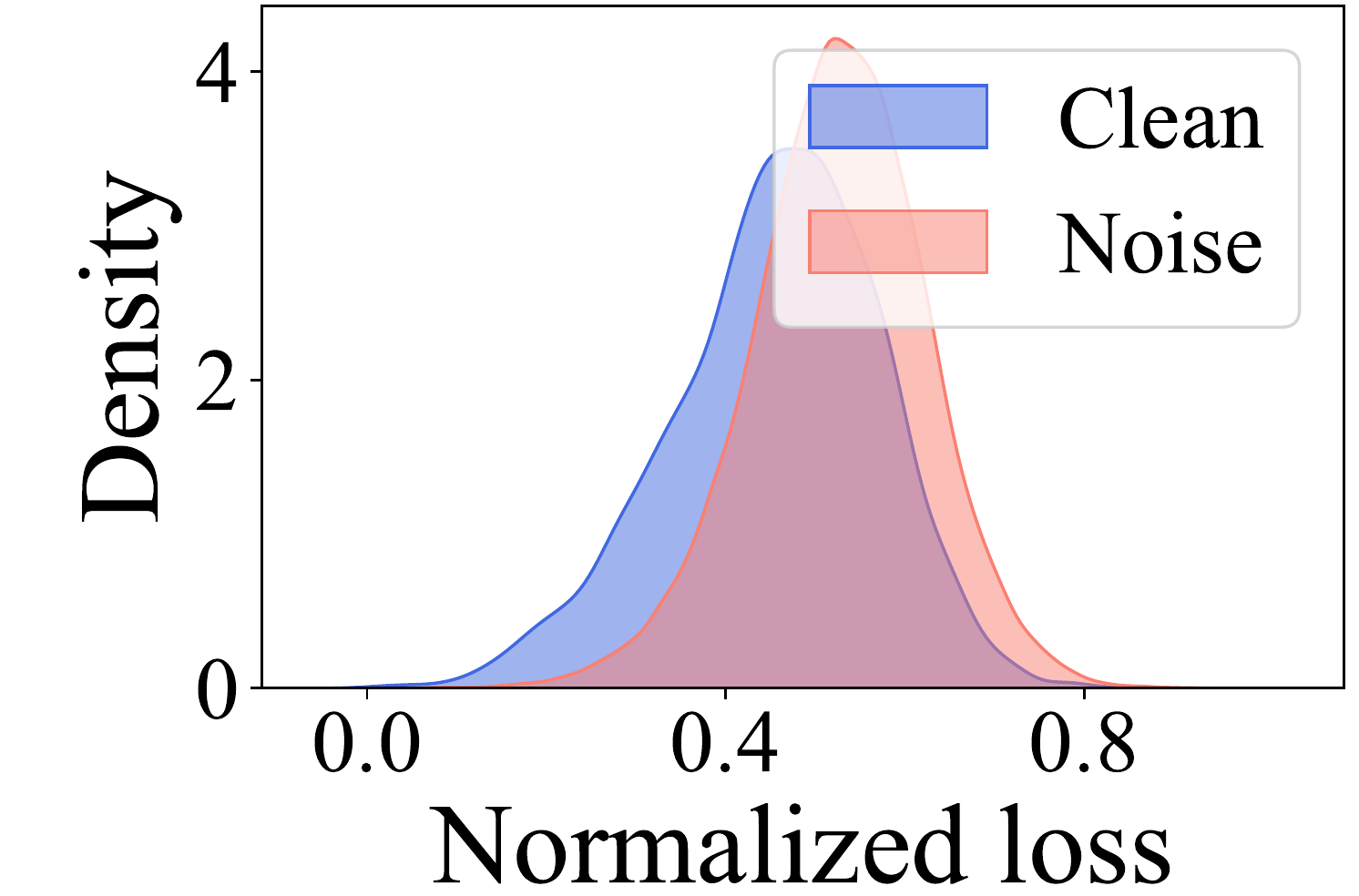}
    \label{fig-90cross}
    }
    \hspace{-.1in}
    \subfigure[Proposed warm-up at 0.9]{
    \includegraphics[width=.24\linewidth]{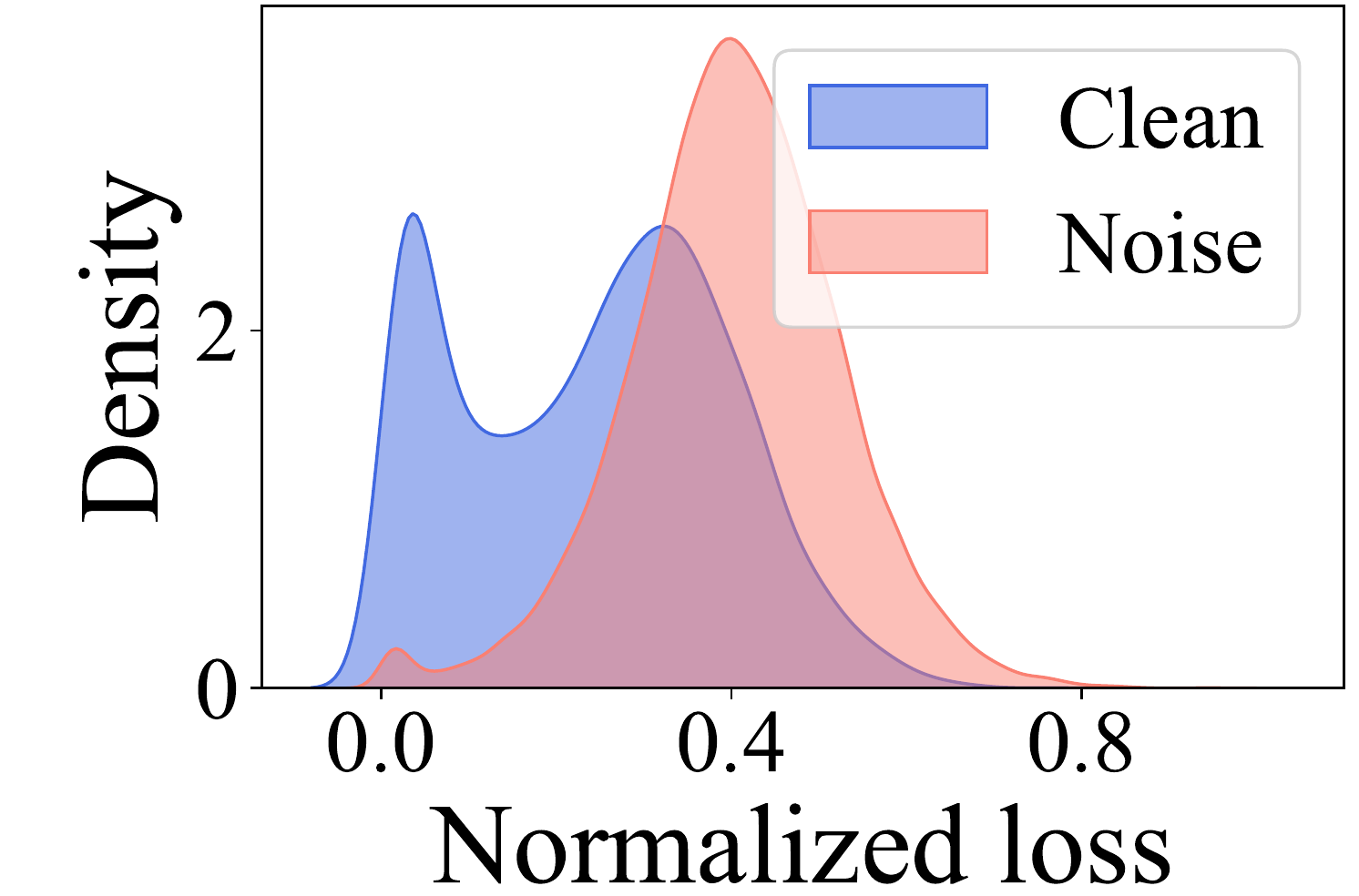}
    \label{fig-90split}
    }
    % \vspace{-10pt}
    \caption{\textbf{Effect of proposed warm-up.} (a), (c), (e), and (g) show the results of warm-up using only cross-entropy. (b), (d), (g), and (h) show the results of our warm-up. With our warm-up, clean and noisy data can be better distinguished.}
    \label{fig:warmupeffect}
    % \vspace{-10pt}
\end{figure*}

\begin{table*}[th!]
  \caption{\textbf{List of hyper-parameters.}}
%   \vspace{-5pt}
  \label{tab:hyperparameters}
  \centering
  \resizebox{0.95\linewidth}{!}{\begin{tabular}{lcccc}
    \toprule
    % & & CIFAR-10 & CIFAR-100 & Clothing1M \\
    & & CIFAR-10 & CIFAR-100 & Food101-N \\
    % \cmidrule(r){1-2} 
    % \cmidrule(l){3-5}
    % Model & Noise & 20\% & 50\% & 80\% \\
    
    % \cmidrule(r){1-1} \cmidrule(rl){2-2} \cmidrule(rl){3-7} \cmidrule(l){8-11}
    \midrule
    \multirow{2}[0]{*}{\textit{K}-fold cross-filtering}  
    % & Model & PreActResNet18~\cite{he2016identity} & PreActResNet18~\cite{he2016identity} & ResNet50~\cite{he2016identity} \\
    & epoch & 20 & 20 & 1 \\
    & \textit{K} & 8 & 8 & 8 \\
    
    \cmidrule(r){1-1} \cmidrule(l){2-5}
    \multirow{2}[0]{*}{Warm-up SSL}  
    % & Model & PreActResNet18~\cite{he2016identity} & PreActResNet18~\cite{he2016identity} & ResNet50~\cite{he2016identity} \\
    & epoch & 30 & 50 & 30 \\
    & mixup ${\alpha}$ & 4 & 4 & 4 \\
    
    \cmidrule(r){1-1} \cmidrule(l){2-5}
    \multirow{9}[0]{*}{Main training} 
    % & Model & PreActResNet18~\cite{he2016identity} & PreActResNet18~\cite{he2016identity} & ResNet50~\cite{he2016identity} \\
    & epoch & 300 & 400 & 300 \\
    & learning rate decay & 1/10 at 200 epoch & 1/10 at 300 epoch & 1/10 at 200epoch \\
    & weight decay & 5e-4 & 5e-4 & 1e-3 \\
    & batch size & 128 & 128 & 64 \\
    & SGD momentum & 0.9 & 0.9 & 0.9 \\
    & learning rate & 0.02 & 0.02 & 0.002 \\
    & ${\beta_1}$ & 0.95 & 0.95 & 1.0 \\
    & ${\beta_2}$ & 0.5 & 0.5 & 0.7 \\
    & ${\tau_{label}}$ & 0.95 & 0.95 & 0.95 \\
    % & main network optimizer & 128 & 128 & 32 \\
    % & SplitNet optimizer & 128 & 128 & 32 \\
    
    \bottomrule
  \end{tabular}}
\end{table*}

%\subsubsection{$K$-Fold Cross Filtering}

% $K$-fold cross-validation is the method most prevalently used to evaluate model performance when data is independent and identically distributed.

Our warm-up consists of $K$-fold cross-filtering and SSL boosting. $K$-fold cross-filtering first divides the data into $K$-folds and then checks whether the labels of the test data match through out-of-fold prediction. By K-fold cross-filtering, we can find noisy data, discard their labels, and warm up the main network using the SSL method presented in~\cite{sohn2020fixmatch}.

% Through this method, we exploit the SSL method presented in~\cite{sohn2020fixmatch}, which utilizes filtered data as labeled dataset to warm-up the main network. 
% Inspired by this, we propose $K$-fold cross-filtering which divides the data into $K$-folds, checks whether the labels of the test data match through out-of-fold prediction, and checks whether the confidence of predicted labels exceeds the threshold. Through this method, we exploit the SSL method presented in~\cite{sohn2020fixmatch}, which utilizes filtered data as labeled dataset to warm-up the main network. 
In DNNs, learning fewer noise samples is important to make a distinguishable loss distribution since the loss value of incorrectly labeled samples quickly decreases to the loss value of correctly labeled samples. The similarity in loss value between correctly and incorrectly labeled samples makes it difficult to distinguish when learning with incorrect labels. Therefore, we select safer samples through $K$-fold cross-filtering to maintain a high loss value for noise samples. The higher the noise ratio, the greater the effect because more noise samples are removed.

Formally, when training dataset $\mathcal{X}$ is divided into a number $\mathcal{K}$ of folds of equal size, let the $k$-th fold be $\mathcal{X}^k$. $\theta^k_f$ refers to filtering network parameters that are trained by cross-entropy loss on $\mathcal{O^\mathnormal{k}}=\mathcal{X}-\mathcal{X}^k$. It follows that $\theta^k_f$ is trained by the following loss function: 
\begin{equation}\label{ltheta}
    \ell(\theta^k_f) = -\frac{1}{\mid\mathcal{O^\mathnormal{k}}\mid}\sum_{x,y\in\mathcal{O^\mathnormal{k}}}y\,\mathrm{log}(p_f(x;\theta^k_f)).
\end{equation}
where $p_f(x;\theta_f)$ is the filtering network’s predicted class distribution with parameters $\theta_f$. In this case, we define $\mathcal{T}^k$ as the set of data presumed to be clean within $\mathcal{X}^k$ as: 
\begin{equation}\label{tauk}
\begin{split}
   \mathcal{T}^{\,k}=\{(x,y)\mid\mathrm{arg\,max}(y)=\mathrm{arg\,max}(p_\mathnormal{f}(x;\theta^k_f))\\
   \land\mathrm{max}(p_\mathnormal{f}(x;\theta^k_f))\ge\tau_{\mathrm{\ label}}\land(x,y)\in\mathcal{X}^k\}.
\end{split}
\end{equation}
Then, it follows that the clean dataset $\mathcal{T}$ ultimately yielded by $K$-fold cross-filtering can be expressed such that $\mathcal{T}=\{\mathcal{T}^{\,{1}} \cup \mathcal{T}^{\,{2}}\cdot\cdot\cdot\cup\ \mathcal{T}^{\,{\mathcal{K}}}\}$.

Since we configure a clean dataset through the filtering process, our ability to warm-up the main model is enhanced compared to that of the normal cross-entropy. Through the method presented in~\cite{zhang2017mixup}, we train the main model using supervised loss on clean dataset $\mathcal{T}$. Additionally, by utilizing consistency regulation as previously presented in ~\cite{sohn2020fixmatch}, we train the model on all data included in the training dataset $\mathcal{X}$. As demonstrated in~\figref{fig:warmupeffect} the loss distribution is more evidently differentiated when our method is used to warm-up.

\begin{table*}[t]
  \caption{\textbf{Performance comparison for our method and the state-of-the-art methods on CIFAR-10 and CIFAR-100.}}
  \small
  \label{tab:main}
  \centering
  \resizebox{\linewidth}{!}{
  \begin{tabular}{llccccccccc}
    \toprule
    & & \multicolumn{5}{c}{CIFAR-10} & \multicolumn{4}{c}{CIFAR-100} \\
    \cmidrule(r){3-7} \cmidrule(l){8-11}
    Model & Noise & 20\% & 50\% & 80\% & 90\% & 40\% Asym & 20\% & 50\% & 80\% & 90\% \\
    
    % \cmidrule(r){1-1} \cmidrule(rl){2-2} \cmidrule(rl){3-7} \cmidrule(l){8-11}
    \midrule
    \multirow{2}[0]{*}{Cross-Entropy}  
    & Best  & 86.8 & 79.4 & 62.9 & 42.7 & - & 62.0 & 46.7 & 19.9 & 10.1 \\
    & Last & 82.7 & 57.9 & 26.1 & 16.8 & - & 61.8 & 37.3 & 8.8 & 3.5 \\
    
    \cmidrule(r){1-1} \cmidrule(l){2-11}
    \multirow{2}[0]{*}{Bootstrap~\cite{reed2014training}} 
    & Best & 86.8 & 79.8 & 63.3 & 42.9 & - & 62.1 & 46.6 & 19.9 & 10.2 \\
    & Last & 82.9 & 58.4 & 26.8 & 17.0 & - & 62.0 & 37.9 & 8.9 & 3.8 \\
    
    \cmidrule(r){1-1} \cmidrule(l){2-11}
    \multirow{2}[0]{*}{F-correction~\cite{patrini2017making}} 
    & Best & 86.8 & 79.8 & 63.3 & 42.9 & 87.2 & 61.5 & 46.6 & 19.9 & 10.2 \\
    & Last & 83.1 & 59.4 & 26.2 & 18.8 & 83.1 & 61.4 & 37.3 & 9.0 & 3.4 \\
    
    \cmidrule(r){1-1} \cmidrule(l){2-11}
    \multirow{2}[0]{*}{Co-teaching${+}$~\cite{yu2019does}} 
    & Best & 89.5 & 85.7 & 67.4 & 47.9 & - & 65.6 & 51.8 & 27.9 & 13.7 \\
    & Last & 88.2 & 84.1 & 45.5 & 30.1 & - & 64.1 & 45.3 & 15.5 & 8.8 \\
    
    \cmidrule(r){1-1} \cmidrule(l){2-11}
    \multirow{2}[0]{*}{Mixup~\cite{zhang2017mixup}} 
    & Best & 95.6 & 87.1 & 71.6 & 52.2 & - & 67.8 & 57.3 & 30.8 & 14.6 \\
    & Last & 92.3 & 77.6 & 46.7 & 43.9 & - & 66.0 & 46.6 & 17.6 & 8.1 \\
    
    \cmidrule(r){1-1} \cmidrule(l){2-11}
    \multirow{2}[0]{*}{P-correction~\cite{yi2019probabilistic}} 
    & Best & 92.4 & 89.1 & 77.5 & 58.9 & 88.5 & 69.4 & 57.5 & 31.1 & 15.3 \\
    & Last & 92.0 & 88.7 & 76.5 & 58.2 & 88.1 & 68.1 & 56.4 & 20.7 & 8.8 \\
    
    \cmidrule(r){1-1} \cmidrule(l){2-11}
    \multirow{2}[0]{*}{Meta-Learning~\cite{li2019learning}} 
    & Best & 92.9 & 89.3 & 77.4 & 58.7 & 89.2 & 68.5 & 59.2 & 42.4 & 19.5 \\
    & Last & 92.0 & 88.8 & 76.1 & 58.3 & 88.6 & 67.7 & 58.0 & 40.1 & 14.3 \\
    
    \cmidrule(r){1-1} \cmidrule(l){2-11}
    \multirow{2}[0]{*}{M-correction~\cite{arazo2019unsupervised}} 
    & Best & 94.0 & 92.0 & 86.8 & 69.1 & 87.4 & 73.9 & 66.1 & 48.2 & 24.3 \\
    & Last & 93.8 & 91.9 & 86.6 & 68.7 & 86.3 & 73.4 & 65.4 & 47.6 & 20.5 \\
    
    \cmidrule(r){1-1} \cmidrule(l){2-11}
    \multirow{2}[0]{*}{DivideMix~\cite{li2020dividemix}} 
    & Best & 96.1 & 94.6 & 93.2 & 76.0 & 93.4 & 77.3 & 74.6 & 60.2 & 31.5 \\
    & Last & 95.7 & 94.4 & 92.9 & 75.4 & 92.1 & 76.9 & 74.2 & 59.6 & 31.0 \\
    
    \cmidrule(r){1-1} \cmidrule(l){2-11}
    \multirow{2}[0]{*}{DM-AugDesc-WS-SAW~\cite{nishi2021augmentation}}  
    & Best & 96.3 & 95.6 & 93.7 & 35.3 & 94.4 & 79.6 & 77.6 & 61.8 & 17.3 \\
    & Last & 96.2 & 95.4 & 93.6 & 10.0 & 94.1 & 79.5 & 77.5 & 61.6 & 15.1 \\
    
    \cmidrule(r){1-1} \cmidrule(l){2-11}
    \multirow{2}[0]{*}{DM-AugDesc-WS-WAW~\cite{nishi2021augmentation}}  
    & Best & 96.3 & 95.4 & 93.8 & 91.9 & 94.6 & 79.5 & 77.2 & 66.4 & 41.2 \\
    & Last & 96.2 & 95.1 & 93.6 & 91.8 & 94.3 & 79.2 & 77.0 & 66.1 & 40.9 \\
    
    \cmidrule(r){1-1} \cmidrule(l){2-11}
    \multirow{2}[0]{*}{Ours}  
    & Best & \textbf{96.5} & \textbf{96.3} & \textbf{95.2} & \textbf{94.0} & \textbf{95.4} & \textbf{80.6} & \textbf{77.8} & \textbf{70.3} & \textbf{50.7} \\
    & Last & \textbf{96.3} & \textbf{96.0} & \textbf{95.0} & \textbf{93.9} & \textbf{95.3} & \textbf{80.3} & \textbf{77.5} & \textbf{70.2} & \textbf{50.4} \\
    
    \bottomrule
  \end{tabular}}
\end{table*}

\begin{table*}[t!]
  \caption{\textbf{Performance comparison for our method and the state-of-the-art methods on CIFAR-10IDN and CIFAR-100IDN.}}
%   \vspace{5pt}
  \label{tab:cifar-idn}
  \small
  \centering
%   \resizebox{\linewidth}{!}{

% {\scriptsize
{
\resizebox{0.8\linewidth}{!}{
  \begin{tabular}{lcccccc}
    \toprule
    & \multicolumn{3}{c}{CIFAR-10IDN} & \multicolumn{3}{c}{CIFAR-100IDN}\\
    \cmidrule(r){2-4} \cmidrule(l){5-7}
    % Model & \textit{0.2} & \textit{0.4} & \textit{0.6} & \textit{0.2} & \textit{0.4} & \textit{0.6} \\
    % Model & $\eta=0.2$ & $\eta=0.4$ & $\eta=0.6$ & $\eta=0.2$ & $\eta=0.4$ & $\eta=0.6$ \\
    Model & 20\% & 40\% & 60\% & 20\% & 40\% & 60\% \\
    
    % \cmidrule(r){1-1} \cmidrule(rl){2-2} \cmidrule(rl){3-7} \cmidrule(l){8-11}
    \midrule
    
    Cross-Entropy&85.45&76.23&59.75&57.79&41.15&25.28\\
    Forward \textit{T}~\cite{patrini2017making}&87.22&79.37&66.56&58.19&42.80&27.91\\
    $L_{\mathrm{DMI}}$~\cite{xu2019l_dmi}&88.57&82.82&69.94&57.90&42.70&26.96\\
    $L_{q}$~\cite{zhang2018generalized}&85.81&74.66&60.76&57.03&39.81&24.87\\
    Co-teaching~\cite{han2018co}&88.87&73.00&62.51&43.30&23.21&12.58\\
    Co-teaching+~\cite{yu2019does}&89.90&73.78&59.22&41.71&24.45&12.58\\
    JoCoR~\cite{wei2020combating}&88.78&71.64&63.46&43.66&23.95&13.16\\
    Reweight-R~\cite{xia2019anchor}&90.04&84.11&72.18&58.00&43.83&36.07\\
    Peer Loss~\cite{liu2020peer}&89.12&83.26&74.53&61.16&47.23&31.71\\
    CORES$^2$~\cite{cheng2020learning}&91.14&83.67&77.68&66.47&58.99&38.55\\
    DivideMix~\cite{li2020dividemix}&93.33&95.07&85.50&79.04&76.08&46.72\\
    CAL~\cite{zhu2021second}&92.01&84.96&79.82&69.11&63.17&43.58\\
    CC~\cite{zhao2022centrality}&93.68&94.97&\textbf{94.95}&79.61&76.58&59.40\\
    \midrule
    Ours&\textbf{96.65}&\textbf{96.18}&94.84&\textbf{80.45}&\textbf{76.97}&\textbf{70.20}\\
    
    \bottomrule
  \end{tabular}}
  }

\end{table*}
% CIFAR-N
% \subsection{Results on CIFAR-10N and CIFAR-100N Benchmarks}

\begin{table*}[t!]
  \caption{\textbf{Performance comparison for our method and the state-of-the-art methods on CIFAR-10N and CIFAR-100N.}}
%   \vspace{5pt}
  \label{tab:cifar-n}
  \centering
%   \resizebox{\linewidth}{!}{

% {\footnotesize
{
\resizebox{\linewidth}{!}{
  \begin{tabular}{lcccccc}
    \toprule
    & \multicolumn{5}{c}{CIFAR-10N} & \multicolumn{1}{c}{CIFAR-100N}\\
    \cmidrule(r){2-6} \cmidrule(l){7-7}
    Model & \textit{Aggregate} & \textit{Random 1} & \textit{Random 2} & \textit{Random 3} & \textit{Worst} & \textit{Noisy} \\
    
    % \cmidrule(r){1-1} \cmidrule(rl){2-2} \cmidrule(rl){3-7} \cmidrule(l){8-11}
    \midrule
    
    Cross-Entropy&87.77&85.02&86.46&85.26&77.69&55.50\\
    Forward \textit{T}~\cite{patrini2017making}&88.24&86.88&86.24&87.04&79.79&57.01\\
    Co-teaching+~\cite{yu2019does}&90.61&89.70&89.47&89.54&83.26&57.88\\
    T-Revision~\cite{xia2019anchor}&88.52&88.33&87.71&87.79&80.48&51.55\\
    Peer Loss~\cite{liu2020peer}&90.75&89.06&88.76&88.57&82.00&57.59\\
    ELR+~\cite{liu2020early}&94.83&94.43&94.20&94.34&91.09&66.72\\
    Positive-LS~\cite{lukasik2020does}&91.57&89.80&89.35&89.82&82.76&55.84\\
    F-Div~\cite{wei2020optimizing}&91.64&89.70&89.79&89.55&82.53&57.10\\
    Divide-Mix~\cite{li2020dividemix}&95.01&95.16&95.23&95.21&92.56&71.13\\
    Negative-LS~\cite{wei2021understanding}&91.97&90.29&90.37&90.13&82.99&58.59\\
    CORES$^{2*}$~\cite{cheng2020learning}&95.25&94.45&94.88&94.74&91.66&55.72\\
    VolMinNet~\cite{li2021provably}&89.70&88.30&88.27&88.19&80.53&57.80\\
    CAL~\cite{zhu2021second}&91.97&90.93&90.75&90.74&85.36&61.73\\
    PES(Semi)~\cite{bai2021understanding}&94.66&95.06&95.19&95.22&92.68&70.36\\
    \midrule
    Ours&\textbf{96.50}&\textbf{96.47}&\textbf{96.42}&\textbf{96.27}&\textbf{94.22}&\textbf{72.61}\\
    
    \bottomrule
  \end{tabular}%}
  }
  }
\end{table*}

\section{Experiments}

In order to evaluate the effectiveness of our method, we conduct experiments on synthetic datasets designed to have a variety of noise ratios and a real-world dataset, all of which follow standard LNL evaluation protocols~\cite{li2020dividemix,nishi2021augmentation,xia2019anchor,wei2021learning}.

\subsection{Experiment Settings}
% We extensively validate our method on three benchmark datasets: CIFAR-10, CIFAR100 (Krizhevsky \& Hinton, 2009), and Clothing1M (Xiao et al., 2015). We train using an 18-layer PreAct Resnet (He et al., 2016) and SGD with a momentum of 0.9, weight reduction of 0.0005, and a batch size of 128. The network is trained for 300 epochs for Cifar10 and 400 epochs for Cifar100. Set the initial learning rate to 0.02 and reduce it by a factor of 10 after 200 epochs. The warm-up period is 50 Epoch for CIFAR-10 and 100 Epoch for CIFAR-100.Clothing1M is a large data set with real noise labels. Clothing1M consists of labels generated from 1 million educational images and surrounding text collected from online shopping websites. We follow our previous work (Li et al., 2019) and use ResNet-50 with ImageNet pretrained weights.
\subsubsection{CIFAR-10 and CIFAR-100}
The CIFAR-10 and CIFAR-100 datasets~\cite{krizhevsky2009learning} each contain 50,000 sizes of 32x32 color images for training. While the CIFAR-10 dataset comprises 10 classes with 6,000 images each, the CIFAR-100 dataset consists of 100 classes with 600 images each. We test two different noise settings, symmetric noise and asymmetric noise~\cite{tanaka2018joint,li2019learning}. In the case of symmetric noise, symmetric noisy labels are developed when the labels of a set proportion of training samples are flipped to other classes’ labels randomly. In the case of asymmetric noise, asymmetric noisy labels are generated by exchanging images from two specific classes with similar characteristics, such as deer to horse.

An 18-layer PreAct Resnet~\cite{he2016identity} is used as the main network and the filtering network, and trained using SGD with momentum of 0.9. We design a SplitNet which consists of three blocks with three layers each (FC layer, batch normalization~\cite{ioffe2015batch}, and ReLU~\cite{agarap2018deep} and one projection layer at the end. For SplitNet training, we use AdamW~\cite{loshchilov2018decoupled} with a weight decay of 0.0005. In order to ensure a fair and objective comparison, the experiment follows the hyper-parameters of the state-of-the-art technique~\cite{nishi2021augmentation}. For all CIFAR experiments, we use the same hyper-parameters  $\beta_{1}=0.5$, $\beta_{2}=0.95$, batch size of 128, and a weight decay of 0.0005 for the main network and filtering network.

A complete list of the utilized hyper-parameters can be found in the~\tabref{tab:hyperparameters}.
% is a list of hyper-parameters that we used. 
Unlike recent studies~\cite{li2020dividemix,nishi2021augmentation}, our method does not need to set a ${\lambda_u}$ that adjusts the weight of the unlabeled loss when training the SSL learner. This is because the weight is adjusted according to the noise ratio by itself through $\eta = \mid\mathcal{C}\mid / \mid\mathcal{X}\mid$.

% A complete list of the utilized hyper-parameters can be found in the~\secref{sec:hyperparam}.
% We test three noise settings: Cifar-S(Synthetic Noise), Cifar-IDN( instance-dependent Noise), and Cifar-N(Human Noise). 

% \paragraph{CIFAR-10IDN and CIFAR-100IDN.}
% - CLASS-DEPENDENT LABEL NOISE. Synthetic noise consists of symmetric noise that randomly shuffles labels and asymmetric noise that shuffles labels between similar classes. 

\subsubsection{CIFAR-IDN}
CIFAR10IDN and CIFAR-100IDN~\cite{xia2019anchor,chen2021beyond} are datasets that have synthetically injected part-dependent label noise into CIFAR-10 and CIFAR 100, respectively. They are derived from the fact that humans perceive instances by breaking them down into parts and estimate the IDN transition matrix of an instance as a combination of the transition matrices of different parts of the instance. Experiment settings, including hyper-parameters, are identical as in the case of CIFAR10 and CIFAR100.

\subsubsection{CIFAR-N}
~\cite{wei2021learning} presents the CIFAR-N dataset consisting of CIFAR-10N and CIFAR-100N. CIFAR-N equips the training datasets of CIFAR-10 and CIFAR-100 with human-annotated real-world noisy labels, which are collected from Amazon Mechanical Turk. Unlike existing real-world noisy datasets, CIFAR-N is a real-world noisy dataset that establishes controllable, easy-to-use, and moderated-sized with both ground-truth and noisy labels. Experiment settings, including hyper-parameters, are identical as in the case of CIFAR10 and CIFAR100.

% We test two different noise settings, symmetric noise and asymmetric noise~\cite{tanaka2018joint,li2019learning}. In case of symmetric noise, symmetric noisy labels are developed when the labels of a set proportion of training samples are flipped to other class’s labels randomly. In case of asymmetric noise, asymmetric noisy labels are generated by exchanging images from two specific classes with similar characteristics such as deer to horse.

% \paragraph{Clothing 1M.}
% Clothing 1M is a large-scale dataset with real-world noisy labels consisting of one million images from online shopping websites allocated in 14 classes. For a fair comparison, we follow the previous work~\cite{li2019learning} and use ResNet-50 with ImageNet~\cite{deng2009imagenet} pretrained weights.

% Food101-N

\subsubsection{Food101N}

Food101N~\cite{lee2018cleannet} is a large-scale dataset with real-world noisy labels consisting of 31k images from online websites allocated in 101 classes. Image classification is evaluated on Food-101~\cite{kaur2017combining} test set. For a fair comparison, we follow the previous work~\cite{lee2018cleannet} and use ResNet-50 with ImageNet~\cite{deng2009imagenet} pre-trained weights.
We observed that the train data included data that should not be learned, which are not included in the given classes in Food101N. For this reason, we set $\beta_1$ and $\beta_2$ as 0.7 and 1.0, each at a high value to mask the excluded data.

\subsection{Experiment Results}
\subsubsection{Results on CIFAR Benchmarks}

As demonstrated in~\tabref{tab:main}, we compare state-of-the-art methods with various ratios of symmetric noise and with 40\% of asymmetric noise. The asymmetric noise is set at 40\% as setting it at a rate higher than 50\% would result in specific classes becoming theoretically indistinguishable~\cite{li2020dividemix}.
We report substantial improvements in performance across all evaluated benchmarks, with the increases in performance becoming even more evident in cases where more challenging strong noise ratios are used. Note that compared to DivideMix~\cite{li2020dividemix} and AugDesc~\cite{nishi2021augmentation}, where well-performing hyper-parameters differ for cases depending on the strength of the noise ratio, and specifically compared to AugDesc, which has separate well-performing models for cases depending on the strength of the noise ratio (i.e., DM-AugDesc-WS-SAW and DM-Aug-Desc-WS-WAW), our method enhances performance using a single model.
~\tabref{tab:cifar-idn} shows that our method outperforms the previous method in 5 settings out of 6 settings on CIFAR-IDN. 
~\tabref{tab:cifar-n} shows that our method achieves state-of-the-art performance in all criteria on CIFAR-N. It should be noted that before our method, previous state-of-the-arts were different in each criterion on the CIFAR-N benchmark.
Note that we use identical hyper-parameters used in all CIFAR experiments.
\subsubsection{Results on Food101N Benchmarks}
\tabref{tab:Food101N} shows the results for Food-101N using validation data on all classes. Our methods show results excelling all existing methods.

\begin{table}[ht!]
  \caption{\textbf{Comparison against previous state-of-the-arts in test accuracy(\%) on Food101-N.}}
%   \vspace{-5pt}
  \label{tab:Food101N}
  \centering
  \resizebox{0.85\linewidth}{!}{
  \begin{tabular}{lcc}
    \toprule
    % \multicolumn{2}{c}{Part}                   \\
    % \cmidrule(r){1-2}
    Model             & Test Accuracy \\
    \midrule
    Standard           & 84.51\\
    CleanNet $\omega_{hard}$~\cite{lee2018cleannet}     & 83.47\\
    CleanNet $\omega_{soft}$~\cite{lee2018cleannet}     & 83.95\\
    DeepSelf~\cite{han2019deep}     & 85.11\\
    Jo-SRC~\cite{yao2021jo}     & 86.66\\
    PNP-hard~\cite{sun2022pnp}     & 87.31\\
    PNP-soft~\cite{sun2022pnp}     & 87.50\\
    \cmidrule(r){1-2}
    Ours                & \textbf{88.29}\\
    \bottomrule
  \end{tabular}}
\end{table}

\begin{figure*}[th!]
    \centering
    \hspace{-.1in}
    \subfigure[F1 Score / 20\% ratio]{
    \includegraphics[width=.24\linewidth]{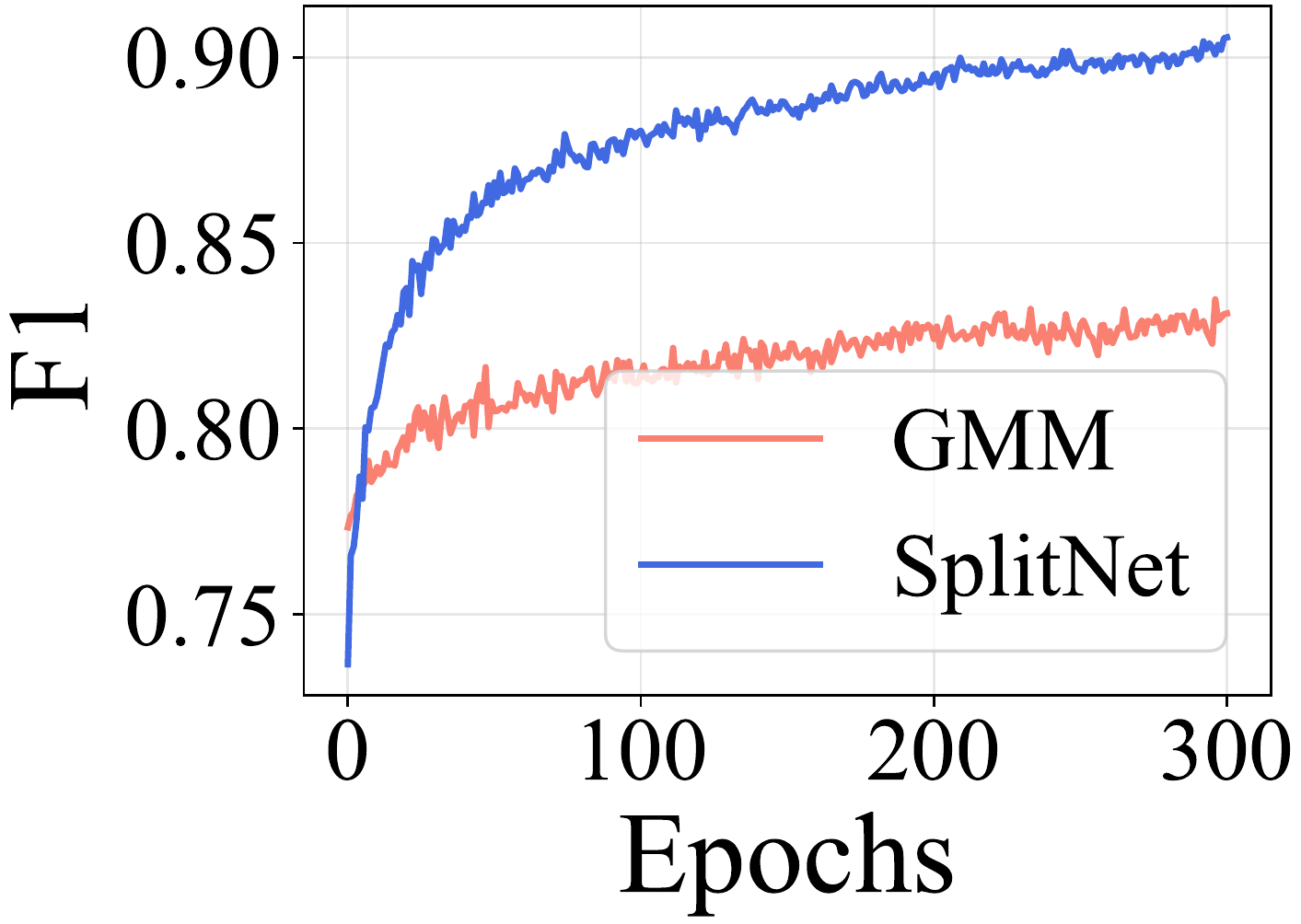}
    \label{fig:20f1supple}
    }
    \hspace{-.1in}
    \subfigure[F1 Score / 50\% ratio]{
    \includegraphics[width=.24\linewidth]{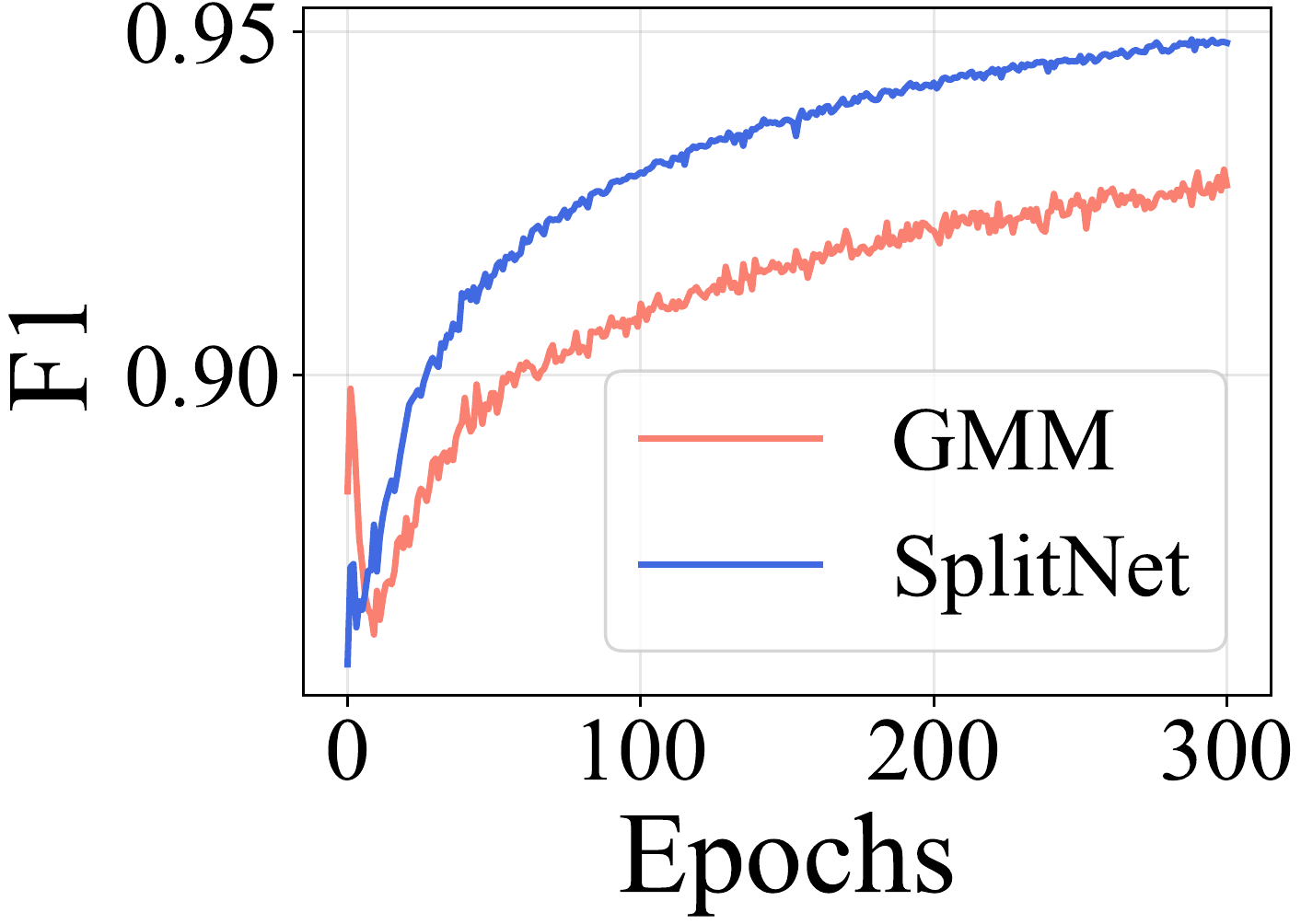}
    \label{fig:50f1supple}
    }
    \hspace{-.1in}
    \subfigure[F1 Score / 80\% ratio]{
    \includegraphics[width=.24\linewidth]{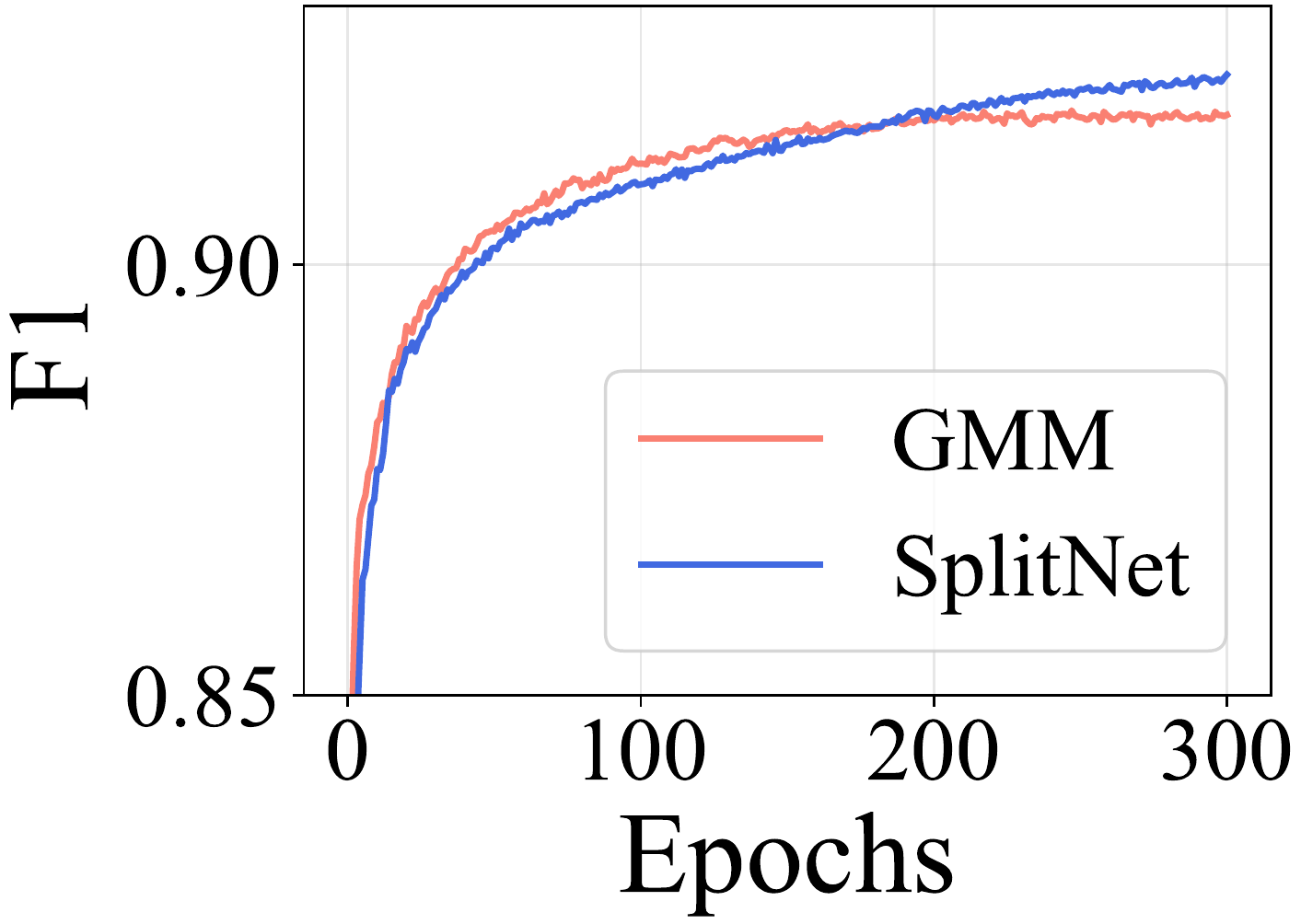}
    \label{fig:80f1supple}
    }
    \hspace{-.1in}
    \subfigure[F1 Score / 90\% ratio]{
    \includegraphics[width=.24\linewidth]{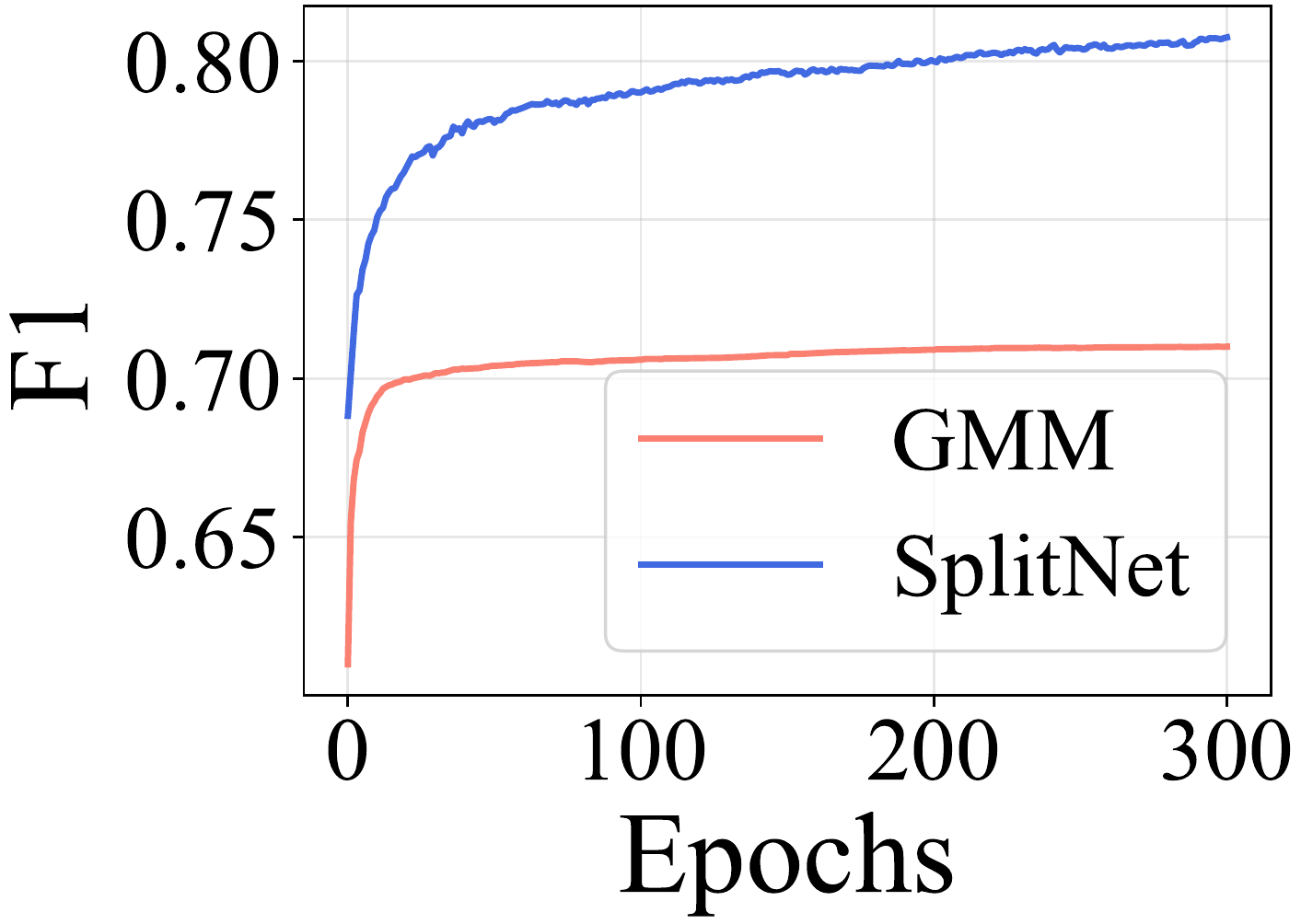}
    \label{fig:90f1supple}
    }\\
    \hspace{-.1in}
    \subfigure[Accuracy / 20\% ratio]{
    \includegraphics[width=.24\linewidth]{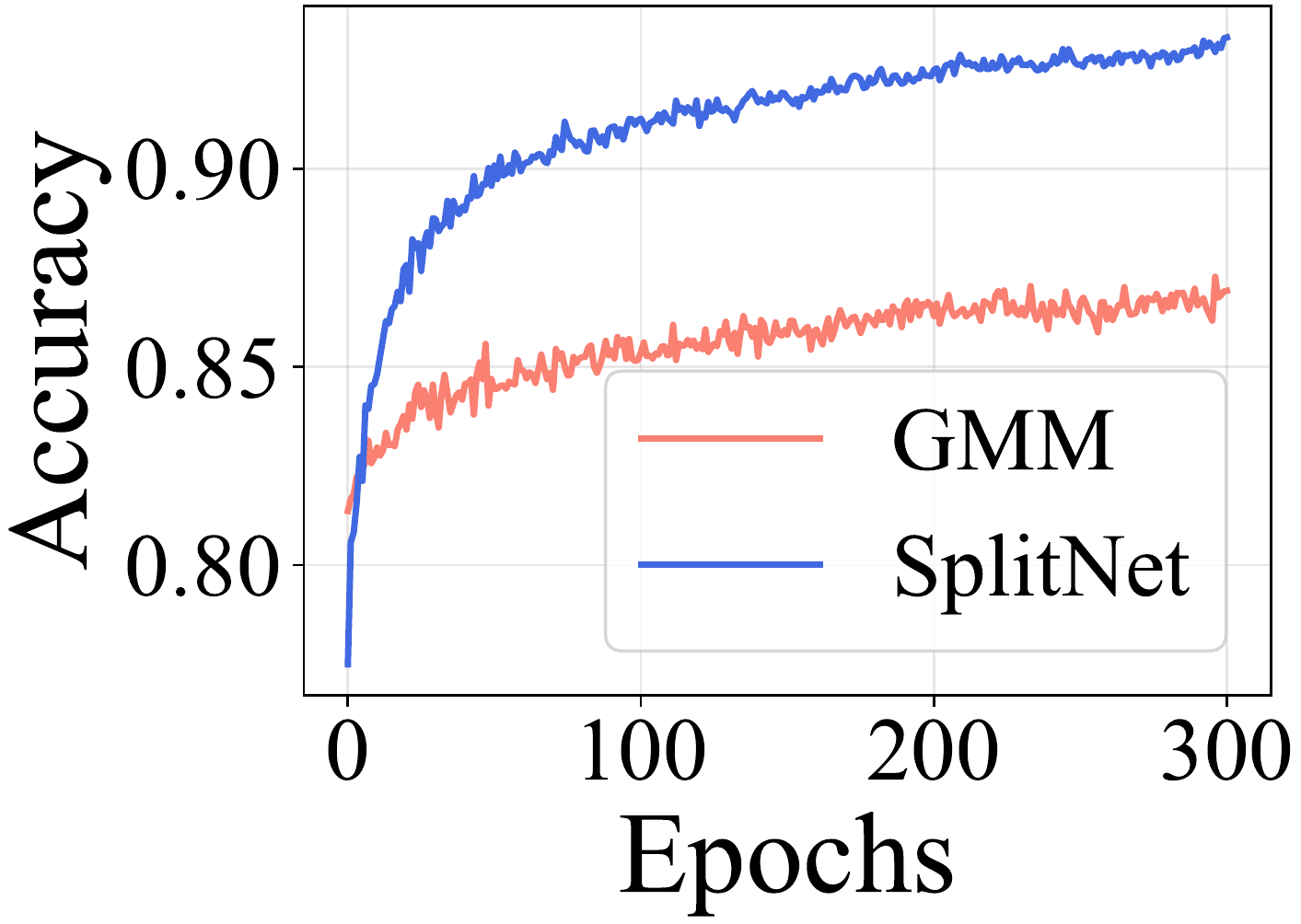}
    \label{fig:20accsupple}
    }
    \hspace{-.1in}
    \subfigure[Accuracy / 50\% ratio]{
    \includegraphics[width=.24\linewidth]{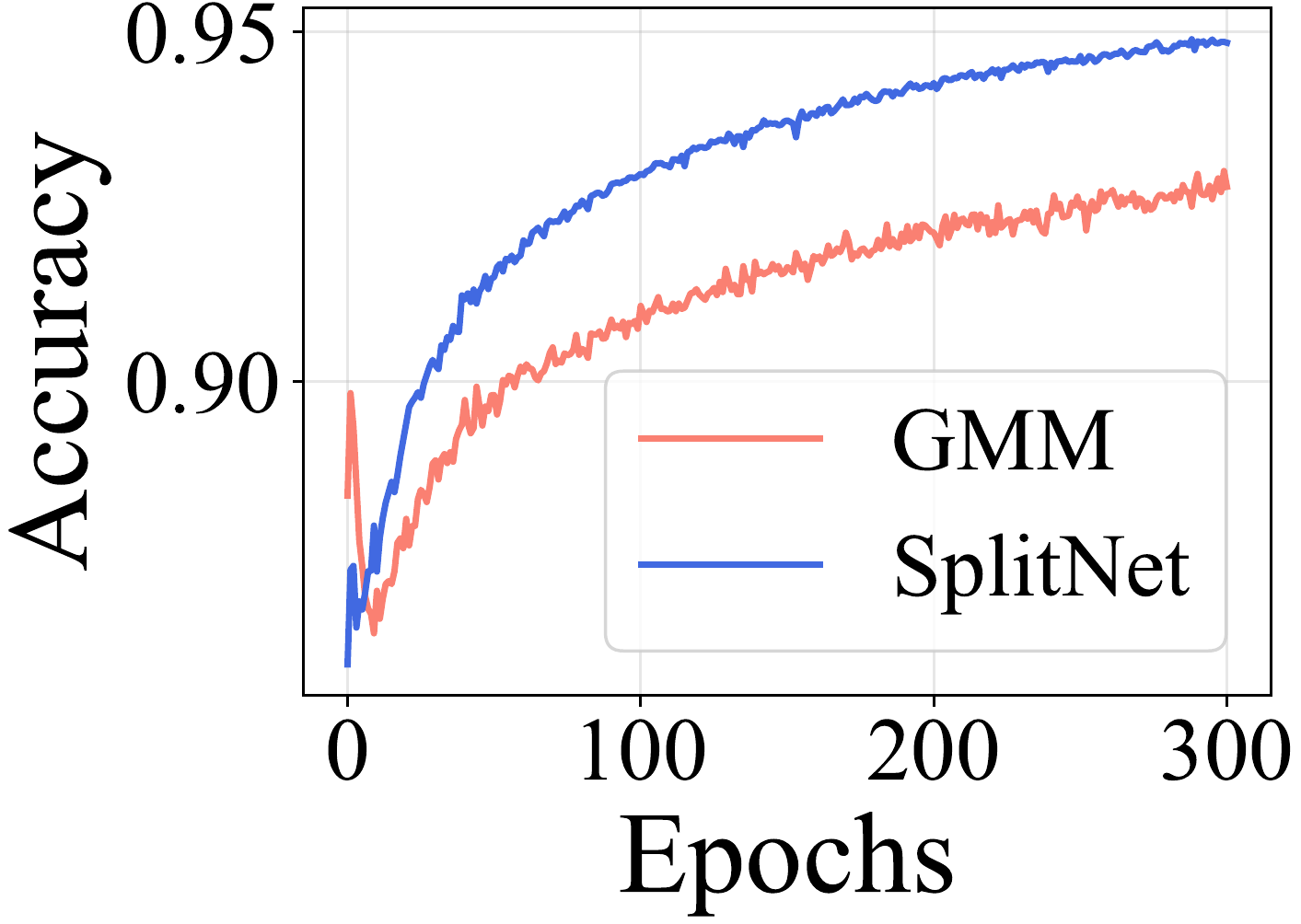}
    \label{fig:50accsupple}
    }
    \hspace{-.1in}
    \subfigure[Accuracy / 80\% ratio]{
    \includegraphics[width=.24\linewidth]{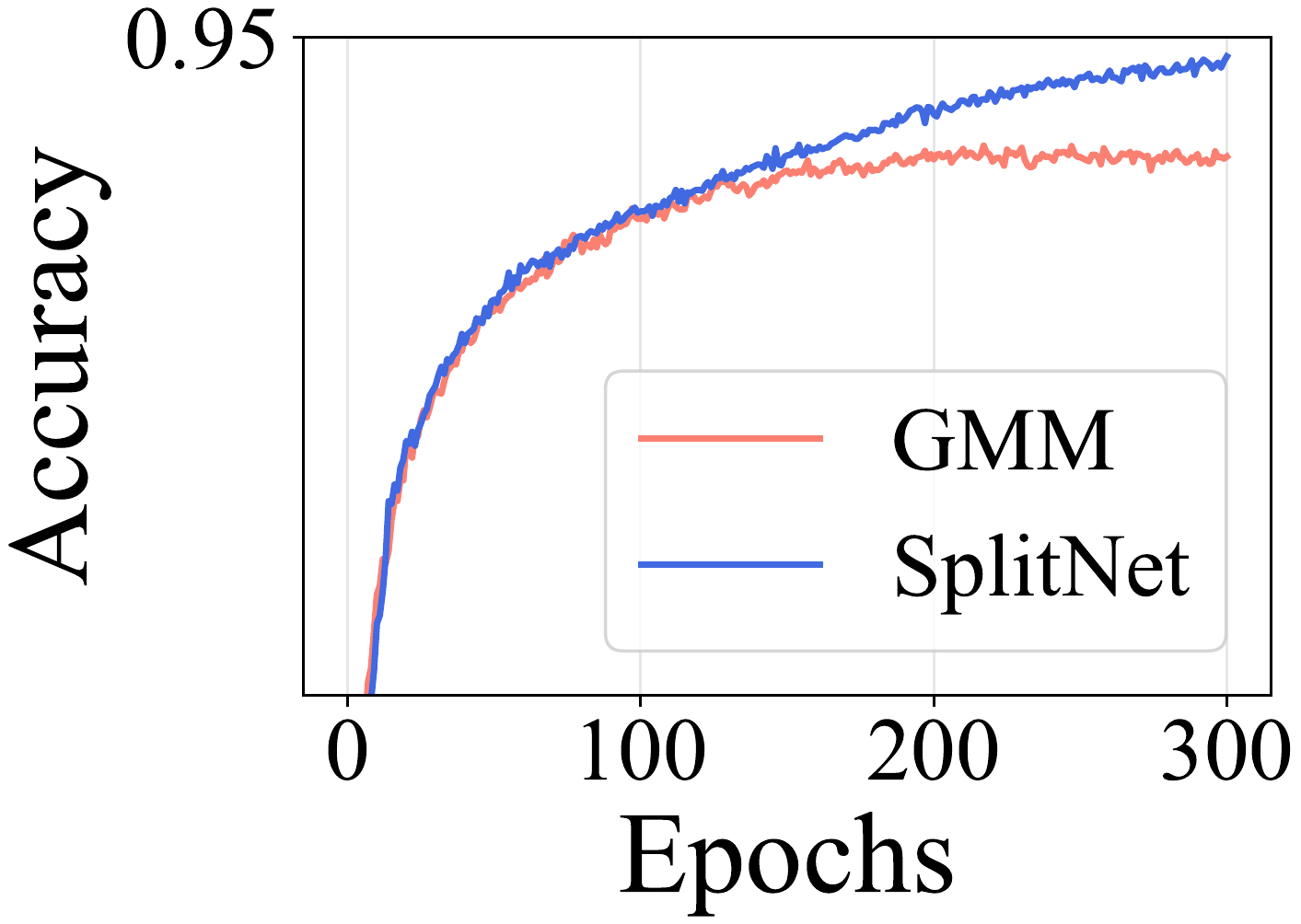}
    \label{fig:80accsupple}
    }
    \hspace{-.1in}
    \subfigure[Accuracy / 90\% ratio]{
    \includegraphics[width=.24\linewidth]{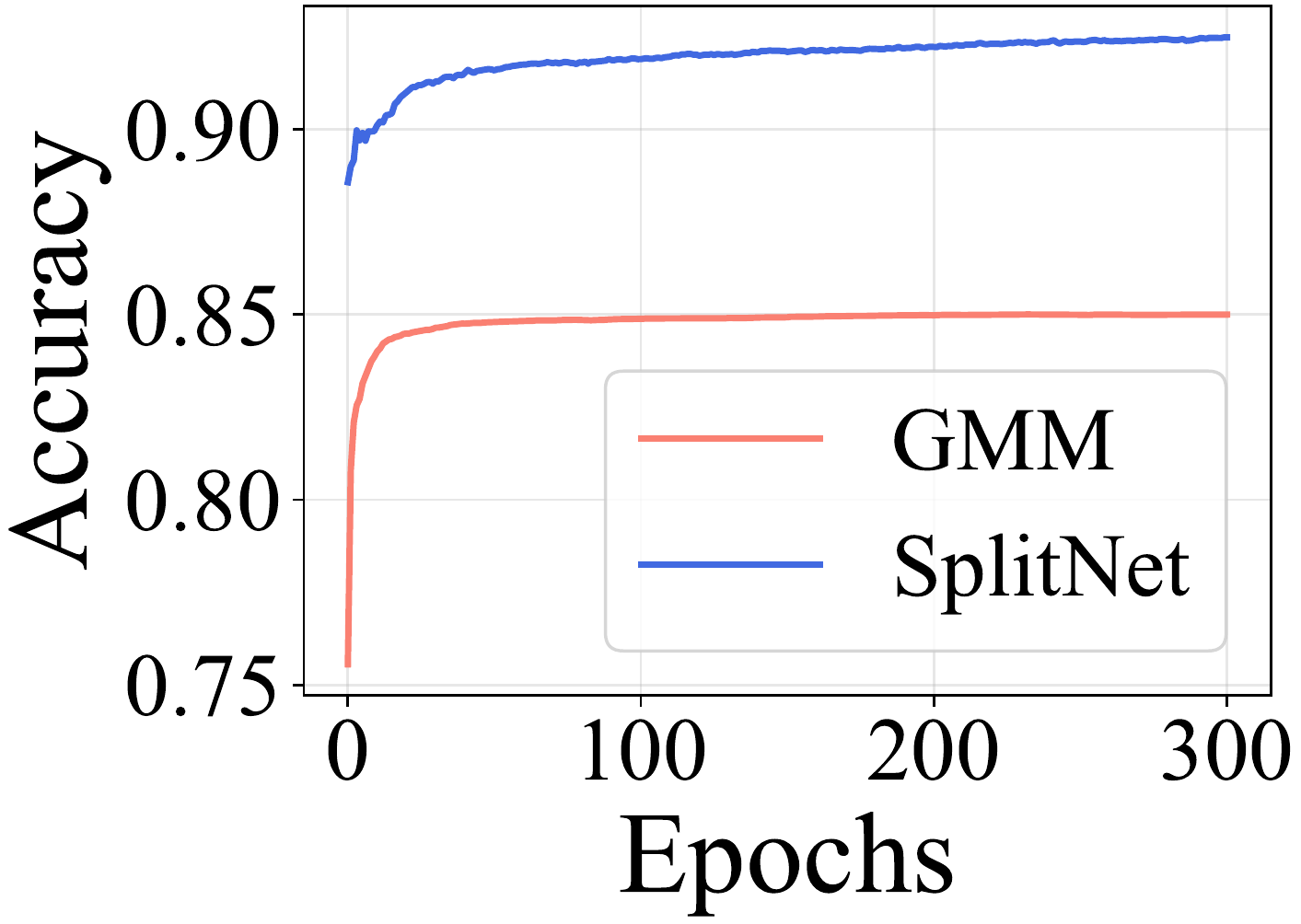}
    \label{fig:90accsupple}
    }
    % \vspace{-5pt}
    \caption{\textbf{Comparison of F1 score and accuracy.} (a), (b), (c), and (d) are the F1 score when the noise ratios are 20\%, 50\%, 80\%, and 90\%, respectively. (e), (f), (g), and (h) are the accuracy when the noise ratios are 20\%, 50\%, 80\%, and 90\%, respectively. For all noise ratios, the F1 score and accuracy of SplitNet are higher, which means that SplitNet selects more actually clean data.}
    % (a) and (e) are the F1 Score and Accuracy when the noise ratio is 20\%, respectively, and (c) and (d) are the F1 Score and Accuracy when the noise ratio is 90\%, respectively.
    \label{fig:lossacc}
    % \vspace{-.1in}
\end{figure*}

\begin{figure*}[th!]
    \centering
    \includegraphics[width=1\linewidth]{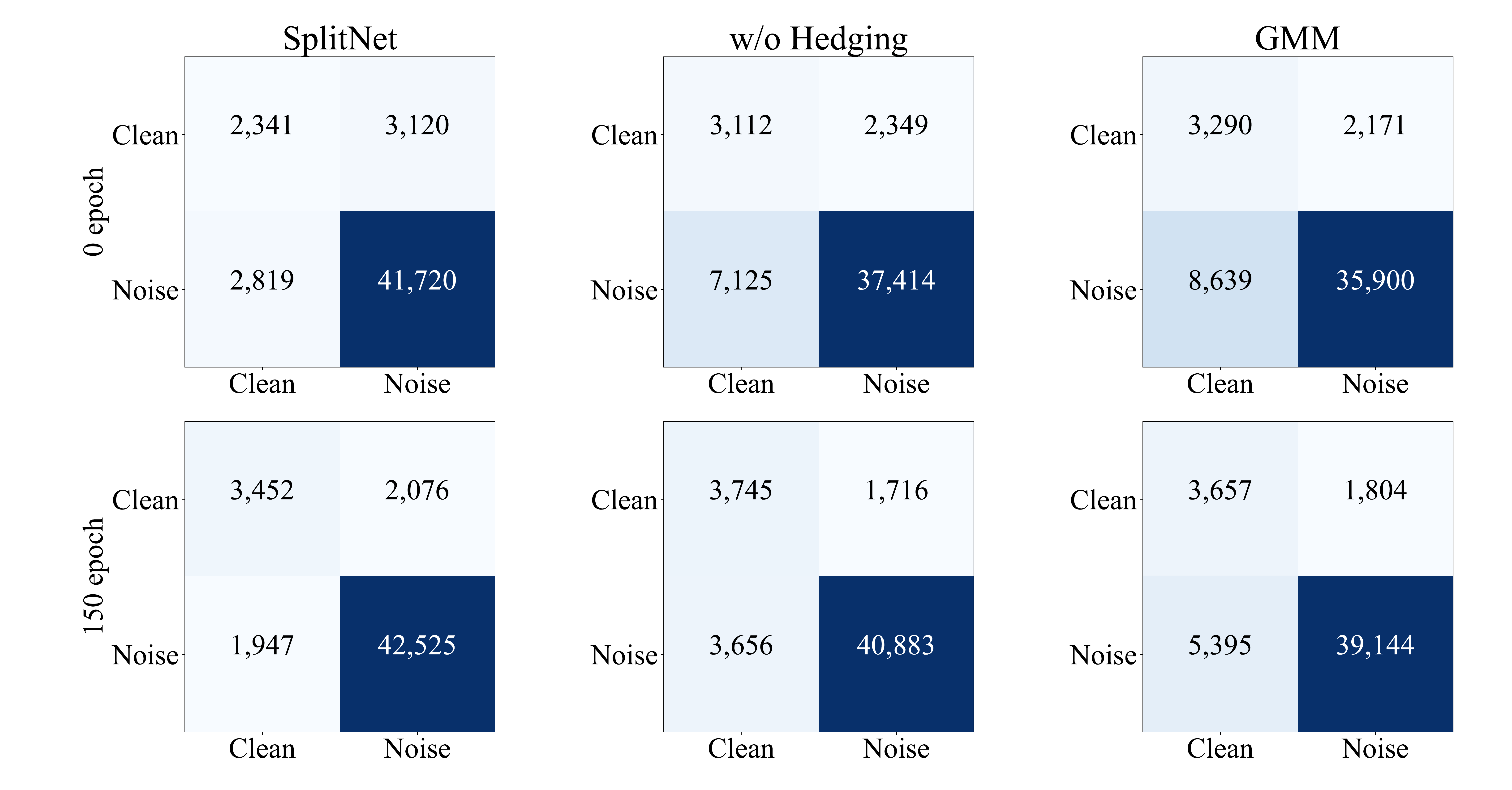}
    \caption{\textbf{Confusion matrix of SplitNet and GMM.} The horizontal axis represents prediction, and the vertical axis represents ground truth in each confusion matrix. The far-left column shows the results of SplitNet trained through hedging after warm-up, the middle column shows the results of SplitNet trained with data filtered with a fixed threshold, and the far-right column shows the results of GMM. The top row shows results at 0 epoch, and the bottom row shows results at 150.}
    \label{fig:confusion}
    % \vspace{-10pt}
\end{figure*}

\section{Analysis}

\subsection{Ablation Study}
% \vspace{-5pt}
% \subsection{Component Analysis}
In order to obtain a better understanding regarding why our method was able to achieve state-of-the-art results, we study the effect of removing certain components.~\tabref{tab:ablation_component} indicates the results obtained when each component is removed. When SplitNet is removed, it becomes a consistency regularization~\cite{sohn2020fixmatch} in general. Also, the absence of the warm-up means that the warm-up is conducted with the existing cross-entropy based method. 
It can be confirmed that SplitNet has a boosting effect on performance across all noise settings and that it is even more effective when used along with the proposed warm-up. 
% In settings with a high noise ratio, the effect of the dynamic threshold is the most important. In contrast, when the noise ratio is not high, the effect of the SplitNet is the most prominent. 
% Unlike the existing state-of-the-art methods~\cite{li2020dividemix,nishi2021augmentation} that train and ensemble two models, we reduce the burden on train resources by ensembling the model of the previous epoch. Moreover, we show excellent performance even without ensemble, as can be seen from the results of \textit{w/o epoch-1 model} in ~\tabref{tab:ablation_component}.

\begin{table}[ht!]
  \caption{\textbf{Ablation study results in terms of test accuracy (\%) on CIFAR-100.}}
  \label{tab:ablation_component}
%   \vspace{-5pt}
  \centering
  \begin{tabular}{lcccc}
    \toprule
    % \multicolumn{2}{c}{Part}                   \\
    % \cmidrule(r){1-2}
    Component             & 20\% & 50\% & 80\% & 90\%\\
    \midrule
    Ours           & \textbf{80.6} & \textbf{77.8} & \textbf{70.3} & \textbf{50.7}\\
    % w/o SplitNet    & 77.7 & 76.9 & 67.2 & 42.4\\
    w/o SplitNet    & 76.8 & 73.2 & 59.1 & 33.4\\
    w/o Warm-up    & 79.8 & 76.7 & 68.8 & 41.0\\
    % w/o Dynamic(0.95)    & 80.2 & 77.4 & 65.3 & 37.3\\
    % w/o Dynamic(0.5)    & 79.9 & 76.7 & 69.2 & 46.5\\
    w/o Hedging    & 79.9 & 77.5 & 69.4 & 46.1\\
    % only fixmatch    & 73.5 & 71.1 & 57.6 & 35.6\\
    divideMix with fixmatch    & 73.8 & 73.5 & 53.8 & 26.5\\
    % w/o epoch-1 model  &80.3 &77.3 &68.6 &50.6  \\
    % \cmidrule(r){1-2}
    % Ours                & 0000 \\
    \bottomrule
  \end{tabular}
\end{table}

% \subsection{SplitNet is very good}

% \subsection{Accuracy and F1 Score of SplitNet}
\subsection{Distinguishing Ability of SplitNet}
\label{sec:accf1}
In this section, we evaluate the F1 score and accuracy of the SplitNet against the conventional method, i.e., GMM. For a more detailed comparison, we also provide a confusion matrix of SplitNet and GMM in~\secref{sec:confusion}.

% \paragraph{Formula of Accuracy and F1 Score Used in The Main Paper}
% \subsubsection{Formula of Accuracy and F1 Score}
\subsubsection{Accuracy and F1 Score}

When selecting clean data, the selected data not only has to be actually clean but \emph{'more actually clean data'} must be selected from the entire dataset to say that it is selected well. Therefore we measured accuracy, a metric that considers the size of the entire dataset. Also, since there is a large difference between the number of clean and noisy data, we verified our method with the f1 score, a metric that takes this into account.

The F1 score and accuracy are defined as follows:

$$
\mathrm{F1\;Score = 2\cdot\frac{precision{\cdot}recall}{precision+recall}}
$$

$$
\mathrm{Accuracy = \cfrac{TP+TN}{TP+FN+FP+TN}}
$$

TP is the number of data predicted to be clean among those that were actually clean, TN is the number of data predicted to be noisy among those that were actually noisy, FN is the number of data predicted to be noisy but was actually clean, and FP is the number of data predicted to be clean but was actually noisy.

\subsubsection{Accuracy and F1 Score Results}
% \subsection{Distinguishing Ability of SplitNet.}
\figref{fig:lossacc} shows the accuracy and F1 score of clean and noise data separated by SplitNet and GMM by epoch. The result shows that SplitNet selects clean data with higher accuracy and F1 score than GMM despite its simple structure regardless of the noise ratio.
% , SplitNet always shows better performance than GMM. 

% \subsection{Confusion Matrix of SplitNet and GMM}
\subsection{Confusion Matrix Comparison}
\label{sec:confusion}

\figref{fig:confusion} shows the performance of SplitNet through a confusion matrix. The experiment was conducted on CIFAR-100 with a noise ratio of 90\%. With SplitNet, the number of False Positives, which are data predicted to be Clean but actually Noisy, decreases dramatically.

% \subsection{SplitNet Architecture}

\subsection{Accuracy According to Structure}
\label{AAS}

\begin{figure}[ht!]
    \centering
    \includegraphics[width=0.9\linewidth]{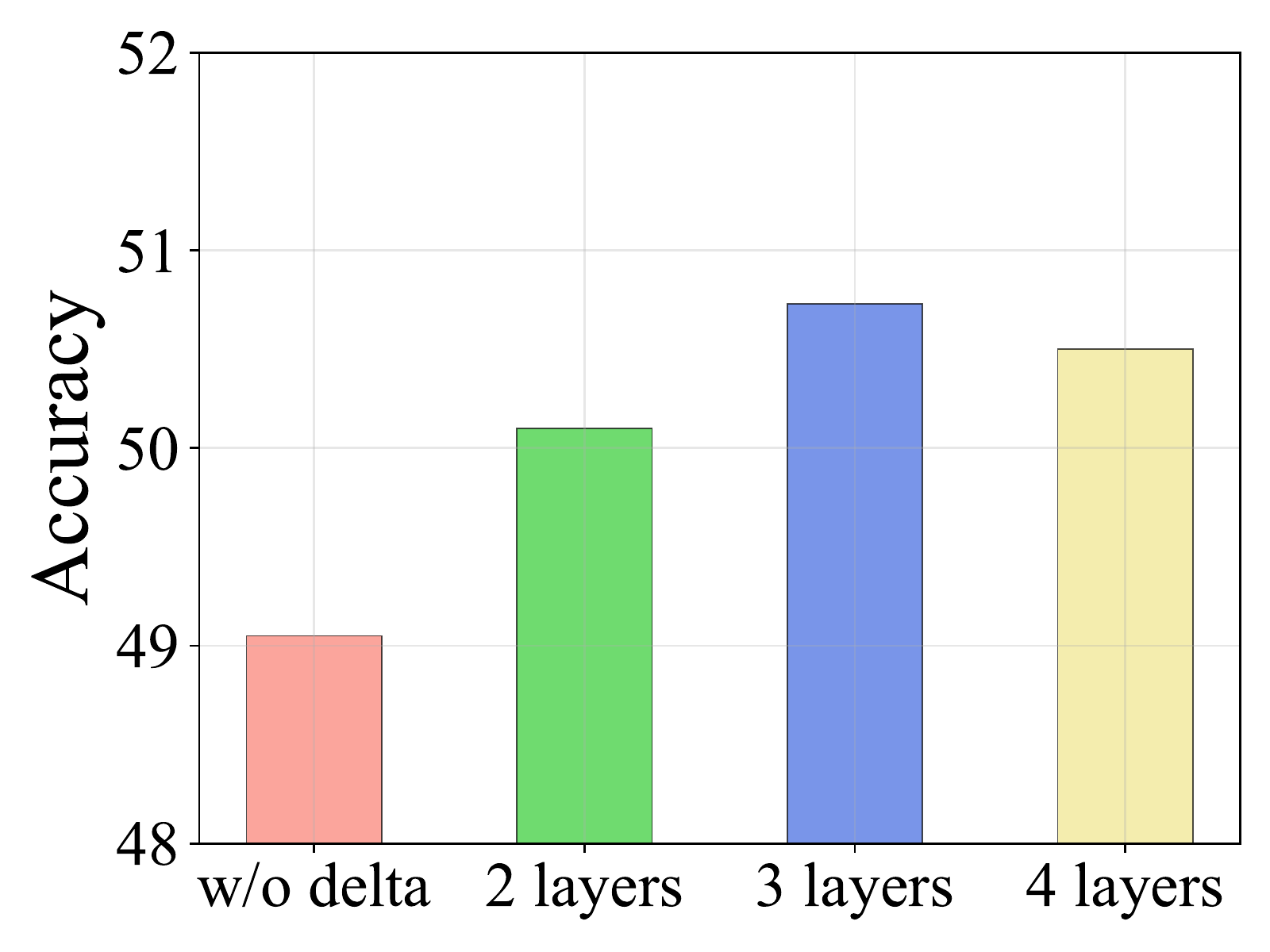}
    \caption{\textbf{Accuracy according to the structure of SplitNet.} w/o delta shows the accuracy when prediction difference is not considered, and 2,3,4 layers show the accuracy when SplitNet is composed of 2,3,4 layers, respectively.}
    \label{fig:splitacc}
\end{figure}

\figref{fig:splitacc} shows the comparison of accuracy according to the structure of SplitNet. It fails to converge when batch normalization is not used (see \figref{fig:nobatchnorm}) and does not show good performance when the prediction difference is not taken into account (see \figref{fig:nodelta}). As shown in \figref{fig:org}, among 2, 3, and 4 layers, it shows the best performance with 3 layers. Future works may try out more various techniques (e.g., residual~\cite{he2016deep}, DenseNet~\cite{huang2017densely}) to improve accuracy.

% \subsection{Dynamic threshold}

% effect of dynamic thresholding by split confidence
\subsection{Effect of Dynamic Thresholding by Split Confidence}
\begin{figure}[ht!]
    \centering
    \hspace{-.1in}
    \subfigure[Number of correct labels.]{
    \includegraphics[width=.48\linewidth]{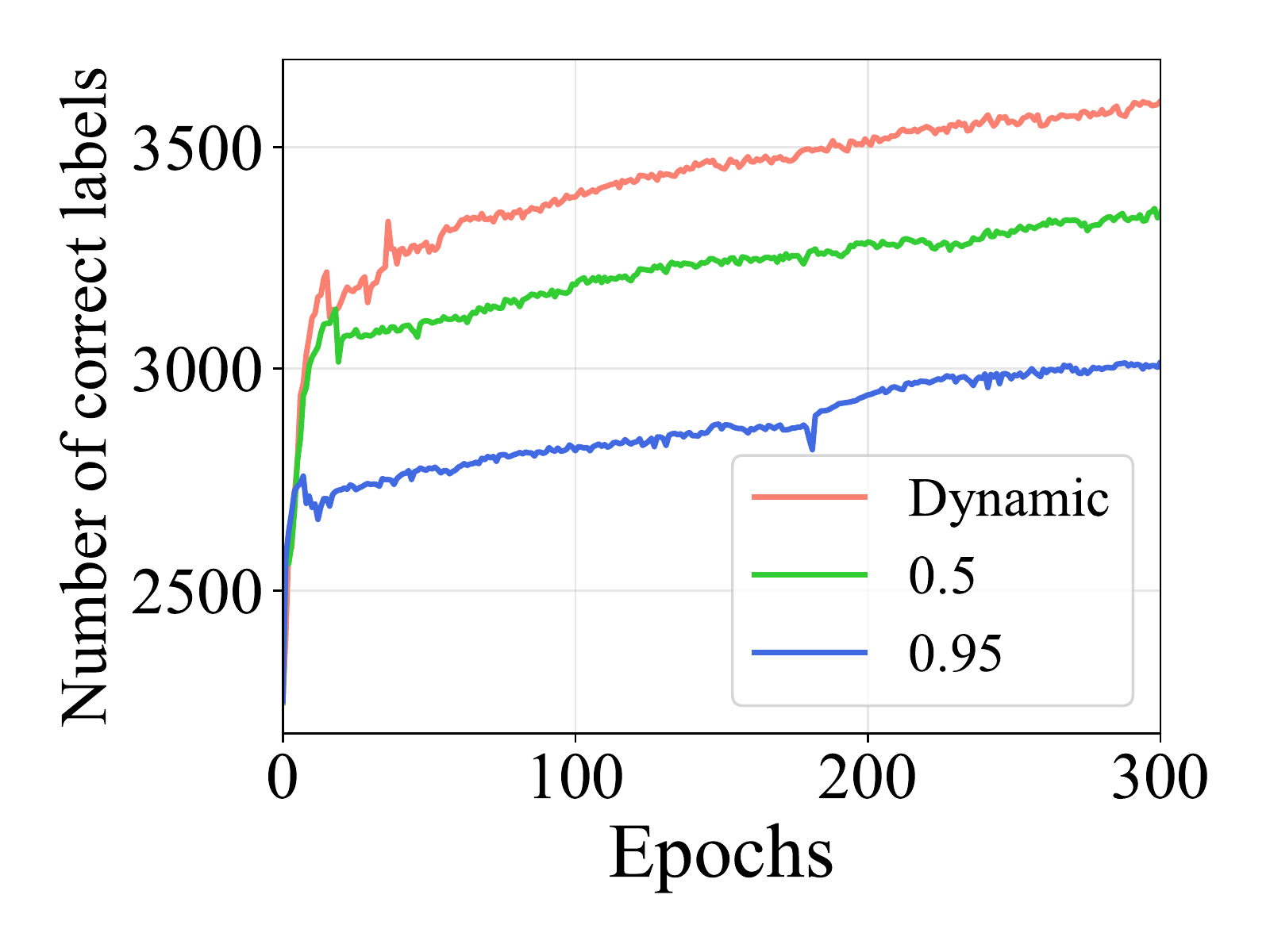}
    \label{fig:correct}
    }
    \hspace{-.1in}
    \subfigure[Number of wrong labels.]{
    \includegraphics[width=.48\linewidth]{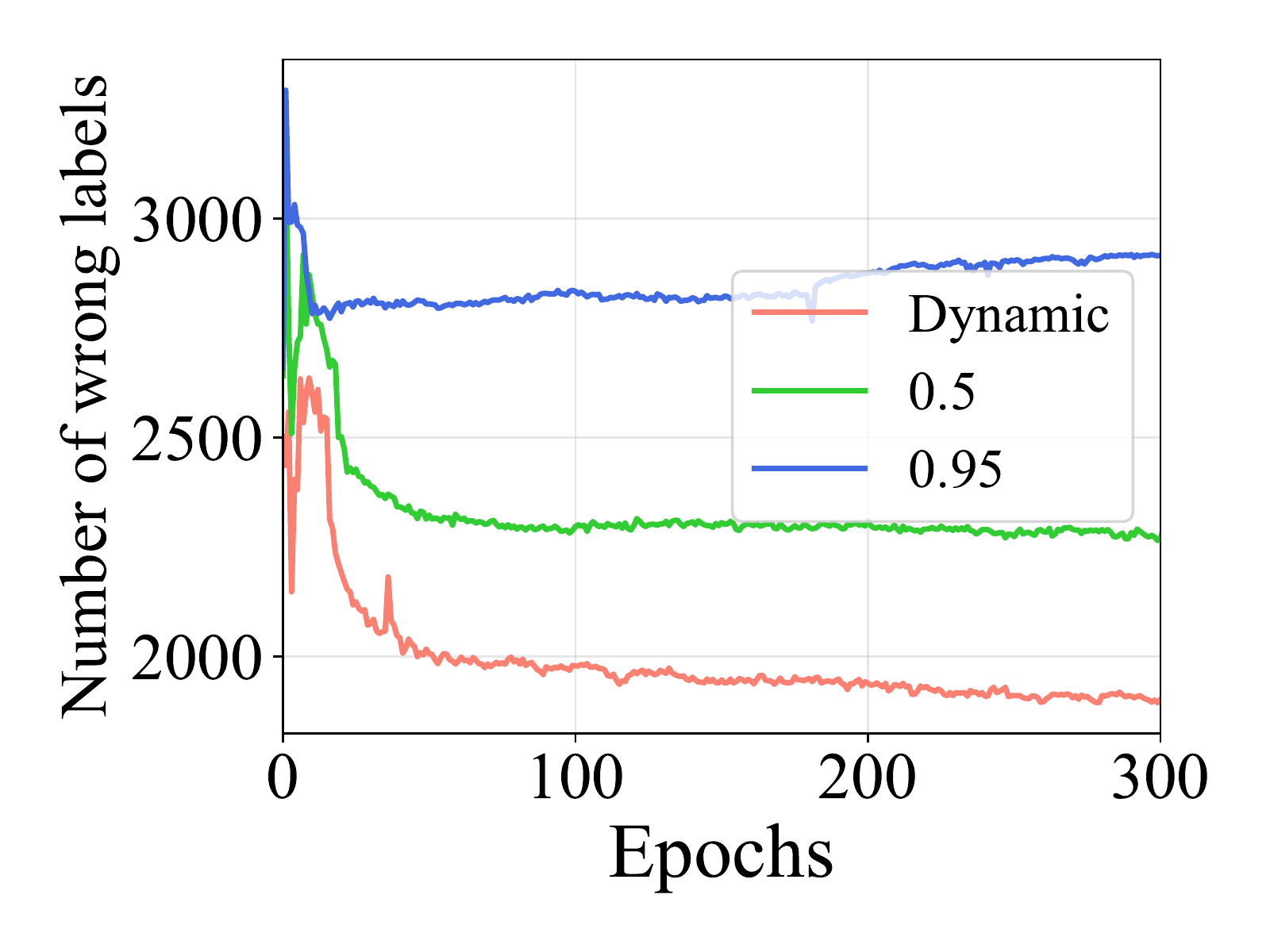}
    \label{fig:wrong}
    }
    % \vspace{-10pt}
    \caption{\textbf{Correctness of pseudo labels by threshold on CIFAR-100 with 90\% noise ratio.} (a) and (b) show the number of correct pseudo labels and wrong pseudo labels, respectively. A dynamic threshold generates more correct pseudo labels and fewer wrong pseudo labels than a fixed threshold.}
    \label{fig:thresholdacc}
    % \vspace{-10pt}
\end{figure}

% As shown in ~\figref{fig:varythreshold}, in the case of a fixed threshold, in situations with a very high level of label noise a lower threshold achieves better performance, and vice versa. This tendency is the reason that achieving superior performance in all noise ratio benchmarks with only one hyper-parameter setting is a difficult task. With motivation from these findings, we propose a dynamic threshold that is adjusted according to the split confidence of the sample. Our dynamic threshold consistently shows higher accuracy on any noise ratio.
%  As shown in ~\figref{fig:varythreshold}, in the case of a fixed threshold, in situations with a very high level of label noise, a lower threshold achieves better performance and vice versa. This tendency is the reason that achieving superior performance in all noise ratio benchmarks with only one hyper-parameter setting is a difficult task. However, our confidence threshold of splitNet consistently shows higher accuracy on any noise ratio. Also, 
When the main network is trained with SSL, pseudo labels are generated for data whose confidence value exceeds the threshold.
As shown in ~\figref{fig:thresholdacc}, the correctness of pseudo labeling is higher with a dynamic threshold as described in equation~\eqref{taudynamic}, compared to when the threshold value is fixed at 0.5 or 0.95.
% \paragraph{Correctness of Pseudo Labels by Threshold.}
% When the main network is trained with SSL, pseudo labels are generated for data whose confidence value exceeds the threshold. As shown in ~\figref{fig:thresholdacc}, the correctness of pseudo labeling is higher with a dynamic threshold, compared to when the threshold value is fixed at 0.5 or 0.95.
%  fixed threshold 를 사용하였을 때 최종적인 

\subsection{\textit{K}-fold Cross-Filtering Performance According to \textit{K}}
\label{sec:Kfold}

\begin{figure}[th!]
    \centering
    \hspace{-.1in}
    \subfigure[80 \% noise ratio]{
    \includegraphics[width=.48\linewidth]{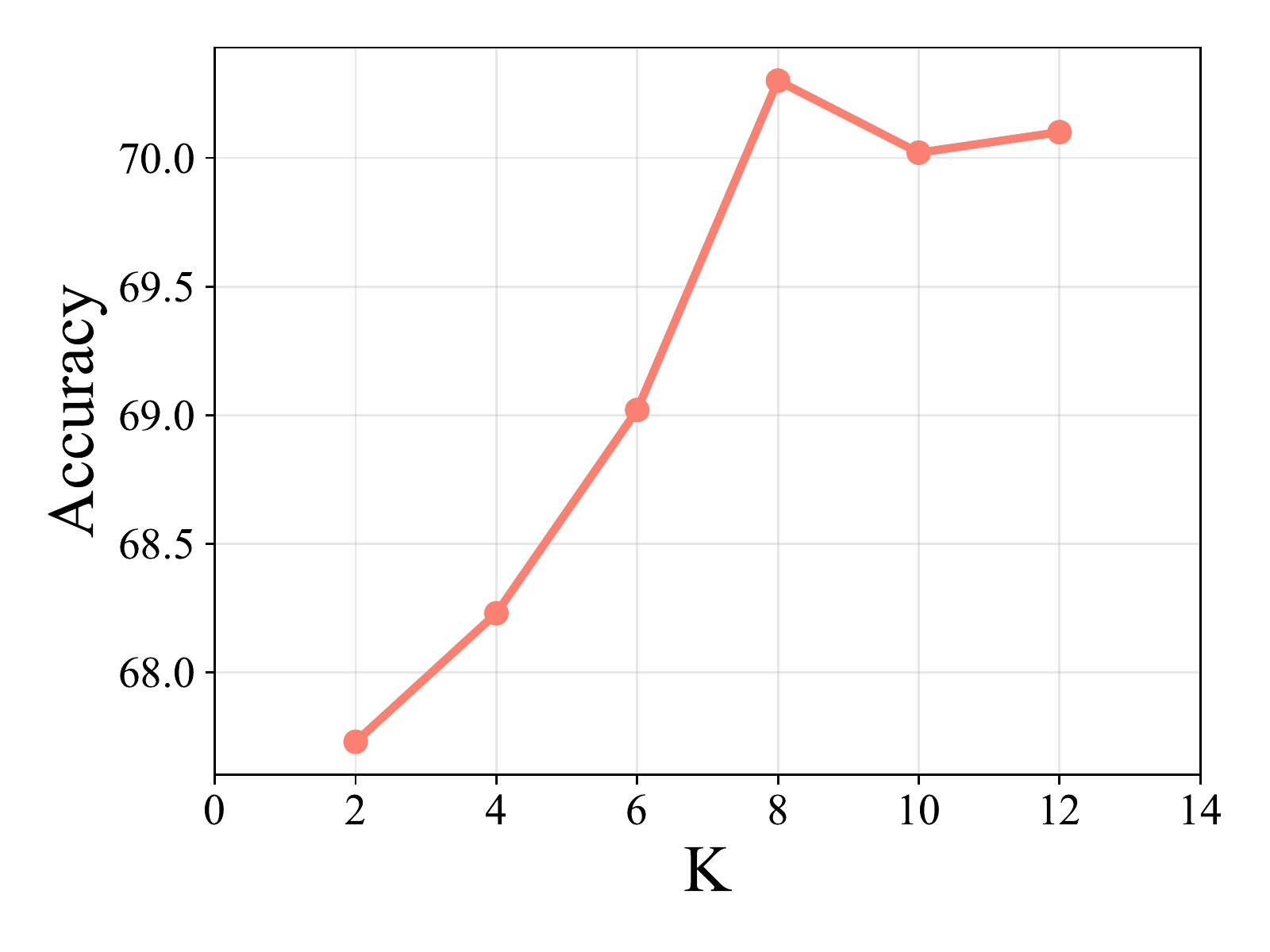}
    \label{fig:80kfold}
    }
    \hspace{-.1in}
    \subfigure[90 \% noise ratio]{
    \includegraphics[width=.48\linewidth]{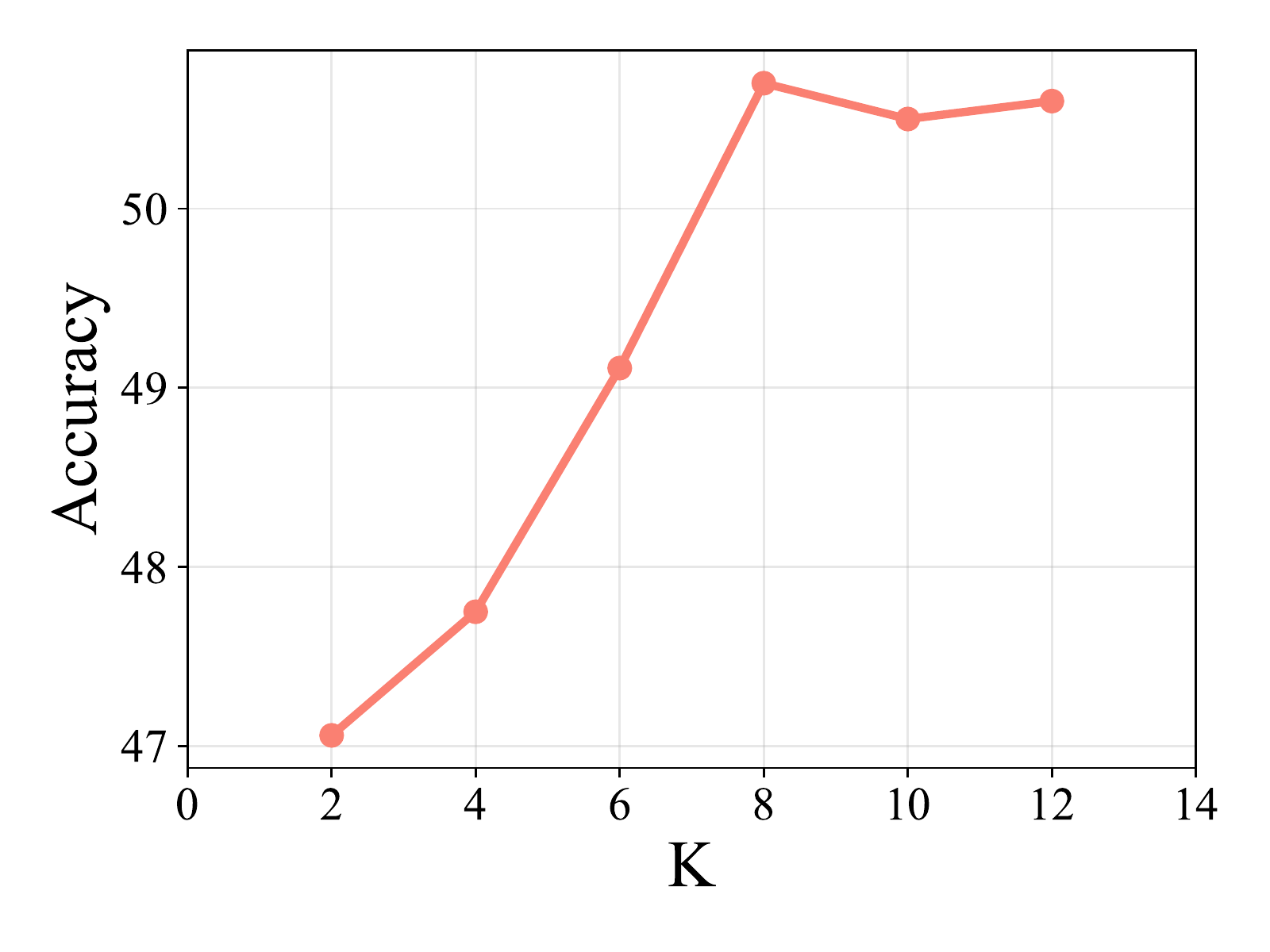}
    \label{fig:90kfold}
    }
    % \vspace{-5pt}
    \caption{\textbf{\textit{K}-fold cross filtering performance evaluation on CIFAR-100 with various noise ratio.}}
    \label{fig:kfoldsupple}
    % \vspace{-.25in}
\end{figure}

\textit{K}, the number of partitions in the dataset in \textit{K}-fold cross-filtering, can be set as a hyper-parameter. \figref{fig:kfoldsupple} shows the test accuracy according to \textit{K}, on noise ratio 80\%, 90\% CIFAR-100. Accuracy is saturated after the value of \textit{K} reaches~8.

\subsection{Training Time Analysis}

As shown in \tabref{tab:time} training time analysis, our method is more efficient compared to conventional methods. 
We compare the training time on CIFAR-10 with 20\% noise ratio.
For a fair comparison, the training times are obtained using a single NVIDIA GeForce RTX 3090 GPU and AMD EPYC 7282 CPU. 
Note that AugDesc~\cite{nishi2021augmentation}, the latest methodology, requires more computational cost, so it requires more training time than DivideMix~\cite{li2020dividemix}.

\begin{table}[ht!]
  \caption{\textbf{Training time comparison.}}
%   \vspace{5pt}
  \label{tab:time}
  \centering
  \resizebox{\linewidth}{!}{\begin{tabular}{lcccc}
    \toprule
    % & & \multicolumn{5}{c}{CIFAR-10} & \multicolumn{4}{c}{CIFAR-100} \\
    % \cmidrule(r){1-3} \cmidrule(l){3-5}
    Model & WarmUp [s] & Main Training [s] & 1 Epoch [s] & Total [h] \\
    
    % \cmidrule(r){1-1} \cmidrule(rl){2-2} \cmidrule(rl){3-7} \cmidrule(l){8-11}
    % \midrule
    \cmidrule(r){1-1} \cmidrule(l){2-5}
    Ours  & 5,271 & 24,420 & 81 & 8.2 \\
    
    % \cmidrule(r){1-1} \cmidrule(l){2-5}
    DivideMix~\cite{li2020dividemix} & 134 & 31,320 & 108 & 8.7 \\
    % AugDesc & ? & ? & ? & ? \\
    
    \bottomrule
  \end{tabular}}
\end{table}

\section{Availability of Supporting Data}
The data that suport the findings of this study are openly available online: 
\begin{enumerate}
\item \textbf{CIFAR~\cite{krizhevsky2009learning}}: \\www.cs.toronto.edu/~kriz/cifar.html
\item \textbf{CIFAR-IDN~\cite{chen2021beyond}}: \\github.com/chenpf1025/IDN
\item \textbf{CIFAR-N~\cite{wei2021learning}}: \\ucsc-real.soe.ucsc.edu:1995/Home.html
\item \textbf{Food101N~\cite{lee2018cleannet}}: \\kuanghuei.github.io/Food-101N/
\end{enumerate}

\section{Conclusion and Discussion}
% \vspace{-8pt}
There has been rapid growth in LNL in recent years. Despite the fact that progress has been accelerated, the setting is becoming more complex due to issues such as having to set different hyper-parameters depending on various noise ratios. The relevance of our method compared to previous ones is that it achieves state-of-the-art performance on most benchmarks with only a single model. Our method enhances the existing warm-up through $K$-fold cross-filtering and SSL Boosting. Additionally, it improves SSL so that it can be better applied to LNL through risk hedging and Dynamic Thresholding. Moreover, we conduct extensive ablation studies to identify why our method is successful and validate the effect of each component. A natural next step, which we leave for future work, is to extend our method to other domains such as audio, text, video, etc.
% \vspace{-10pt}
\paragraph{Broader Impact}
The No Free Lunch Theorem~\cite{wolpert1997no} suggests that there is no single best optimization algorithm~\cite{shalev2014understanding}. Therefore, previous state-of-the-arts used different hyper-parameters depending on the noise ratio according to the noise ratio or created different settings appropriate for each case with different models. Differing from these previous methods, our method designs the model more flexibly so that it can be adapted to various environments. When applying LNL in the real-world, it is not often possible to know the noise ratio of the collected data. Thus, it is very important to study the noise ratio robustly in order to apply LNL to the real environment. The differentiation point of our method is that it can be effectively applied in a real-world environment and can especially be of great aid to organizations with low budgets that face difficulties in obtaining high-quality, refined datasets. 


% \bibliography{sn-bibliography}% common bib file

% \clearpage

\bibliographystyle{spbasic}
\bibliography{egbib} 

\begin{thebibliography}{67}
\providecommand{\natexlab}[1]{#1}
\providecommand{\url}[1]{{#1}}
\providecommand{\urlprefix}{URL }
\expandafter\ifx\csname urlstyle\endcsname\relax
  \providecommand{\doi}[1]{DOI~\discretionary{}{}{}#1}\else
  \providecommand{\doi}{DOI~\discretionary{}{}{}\begingroup
  \urlstyle{rm}\Url}\fi
\providecommand{\eprint}[2][]{\url{#2}}

\bibitem[{Agarap(2018)}]{agarap2018deep}
Agarap AF (2018) Deep learning using rectified linear units (relu). arXiv
  preprint arXiv:180308375

\bibitem[{Arazo et~al(2019)Arazo, Ortego, Albert, O’Connor, and
  McGuinness}]{arazo2019unsupervised}
Arazo E, Ortego D, Albert P, O’Connor N, McGuinness K (2019) Unsupervised
  label noise modeling and loss correction. In: ICML

\bibitem[{Arpit et~al(2017)Arpit, Jastrz{\k{e}}bski, Ballas, Krueger, Bengio,
  Kanwal, Maharaj, Fischer, Courville, Bengio et~al}]{arpit2017closer}
Arpit D, Jastrz{\k{e}}bski S, Ballas N, Krueger D, Bengio E, Kanwal MS, Maharaj
  T, Fischer A, Courville A, Bengio Y, et~al (2017) A closer look at
  memorization in deep networks. In: ICML

\bibitem[{Bai et~al(2021)Bai, Yang, Han, Yang, Li, Mao, Niu, and
  Liu}]{bai2021understanding}
Bai Y, Yang E, Han B, Yang Y, Li J, Mao Y, Niu G, Liu T (2021) Understanding
  and improving early stopping for learning with noisy labels. NeurIPS

\bibitem[{Berthelot et~al(2019{\natexlab{a}})Berthelot, Carlini, Cubuk,
  Kurakin, Sohn, Zhang, and Raffel}]{berthelot2019remixmatch}
Berthelot D, Carlini N, Cubuk ED, Kurakin A, Sohn K, Zhang H, Raffel C
  (2019{\natexlab{a}}) Remixmatch: Semi-supervised learning with distribution
  alignment and augmentation anchoring. arXiv preprint arXiv:191109785

\bibitem[{Berthelot et~al(2019{\natexlab{b}})Berthelot, Carlini, Goodfellow,
  Papernot, Oliver, and Raffel}]{berthelot2019mixmatch}
Berthelot D, Carlini N, Goodfellow I, Papernot N, Oliver A, Raffel CA
  (2019{\natexlab{b}}) Mixmatch: A holistic approach to semi-supervised
  learning. NeurIPS

\bibitem[{Chen et~al(2021{\natexlab{a}})Chen, Cheng, Du, Xu, Jiang, and
  Wang}]{chen2021two}
Chen M, Cheng H, Du Y, Xu M, Jiang W, Wang C (2021{\natexlab{a}}) Two wrongs
  don't make a right: Combating confirmation bias in learning with label noise.
  arXiv preprint arXiv:211202960

\bibitem[{Chen et~al(2021{\natexlab{b}})Chen, Ye, Chen, Zhao, and
  Heng}]{chen2021beyond}
Chen P, Ye J, Chen G, Zhao J, Heng PA (2021{\natexlab{b}}) Beyond
  class-conditional assumption: A primary attempt to combat instance-dependent
  label noise. In: AAAI

\bibitem[{Cheng et~al(2020)Cheng, Zhu, Li, Gong, Sun, and
  Liu}]{cheng2020learning}
Cheng H, Zhu Z, Li X, Gong Y, Sun X, Liu Y (2020) Learning with
  instance-dependent label noise: A sample sieve approach. arXiv preprint
  arXiv:201002347

\bibitem[{Cubuk et~al(2020)Cubuk, Zoph, Shlens, and Le}]{cubuk2020randaugment}
Cubuk ED, Zoph B, Shlens J, Le QV (2020) Randaugment: Practical automated data
  augmentation with a reduced search space. In: CVPR Workshops

\bibitem[{Deng et~al(2009)Deng, Dong, Socher, Li, Li, and
  Fei-Fei}]{deng2009imagenet}
Deng J, Dong W, Socher R, Li LJ, Li K, Fei-Fei L (2009) Imagenet: A large-scale
  hierarchical image database. In: CVPR

\bibitem[{Ding et~al(2018)Ding, Wang, Fan, and Gong}]{ding2018semi}
Ding Y, Wang L, Fan D, Gong B (2018) A semi-supervised two-stage approach to
  learning from noisy labels. In: WACV

\bibitem[{Grandvalet and Bengio(2004)}]{grandvalet2004semi}
Grandvalet Y, Bengio Y (2004) Semi-supervised learning by entropy minimization.
  NeurIPS

\bibitem[{Han et~al(2018)Han, Yao, Yu, Niu, Xu, Hu, Tsang, and
  Sugiyama}]{han2018co}
Han B, Yao Q, Yu X, Niu G, Xu M, Hu W, Tsang I, Sugiyama M (2018) Co-teaching:
  Robust training of deep neural networks with extremely noisy labels. NeurIPS

\bibitem[{Han et~al(2019)Han, Luo, and Wang}]{han2019deep}
Han J, Luo P, Wang X (2019) Deep self-learning from noisy labels. In: ICCV

\bibitem[{He et~al(2016{\natexlab{a}})He, Zhang, Ren, and Sun}]{he2016deep}
He K, Zhang X, Ren S, Sun J (2016{\natexlab{a}}) Deep residual learning for
  image recognition. In: CVPR

\bibitem[{He et~al(2016{\natexlab{b}})He, Zhang, Ren, and Sun}]{he2016identity}
He K, Zhang X, Ren S, Sun J (2016{\natexlab{b}}) Identity mappings in deep
  residual networks. In: ECCV, Springer

\bibitem[{Huang et~al(2017)Huang, Liu, Van Der~Maaten, and
  Weinberger}]{huang2017densely}
Huang G, Liu Z, Van Der~Maaten L, Weinberger KQ (2017) Densely connected
  convolutional networks. In: CVPR

\bibitem[{Ioffe and Szegedy(2015)}]{ioffe2015batch}
Ioffe S, Szegedy C (2015) Batch normalization: Accelerating deep network
  training by reducing internal covariate shift. In: ICML

\bibitem[{Iscen et~al(2022)Iscen, Valmadre, Arnab, and
  Schmid}]{iscen2022learning}
Iscen A, Valmadre J, Arnab A, Schmid C (2022) Learning with neighbor
  consistency for noisy labels. In: CVPR

\bibitem[{Kaur et~al(2017)Kaur, Sikka, and Divakaran}]{kaur2017combining}
Kaur P, Sikka K, Divakaran A (2017) Combining weakly and webly supervised
  learning for classifying food images. arXiv preprint arXiv:171208730

\bibitem[{Kong et~al(2019)Kong, Lee, Kwak, Kang, Kim, and
  Song}]{kong2019recycling}
Kong K, Lee J, Kwak Y, Kang M, Kim SG, Song WJ (2019) Recycling:
  Semi-supervised learning with noisy labels in deep neural networks. Access

\bibitem[{Krizhevsky(2009)}]{krizhevsky2009learning}
Krizhevsky A (2009) Learning multiple layers of features from tiny images.
  Master's thesis, University of Tront

\bibitem[{Krizhevsky et~al(2012)Krizhevsky, Sutskever, and
  Hinton}]{krizhevsky2012imagenet}
Krizhevsky A, Sutskever I, Hinton GE (2012) Imagenet classification with deep
  convolutional neural networks. NeurIPS

\bibitem[{Laine and Aila(2016)}]{laine2016temporal}
Laine S, Aila T (2016) Temporal ensembling for semi-supervised learning. arXiv
  preprint arXiv:161002242

\bibitem[{Lee et~al(2013)}]{lee2013pseudo}
Lee DH, et~al (2013) Pseudo-label: The simple and efficient semi-supervised
  learning method for deep neural networks. In: ICML Workshop

\bibitem[{Lee et~al(2018)Lee, He, Zhang, and Yang}]{lee2018cleannet}
Lee KH, He X, Zhang L, Yang L (2018) Cleannet: Transfer learning for scalable
  image classifier training with label noise. In: CVPR

\bibitem[{Li et~al(2019)Li, Wong, Zhao, and Kankanhalli}]{li2019learning}
Li J, Wong Y, Zhao Q, Kankanhalli MS (2019) Learning to learn from noisy
  labeled data. In: CVPR

\bibitem[{Li et~al(2020)Li, Socher, and Hoi}]{li2020dividemix}
Li J, Socher R, Hoi SC (2020) Dividemix: Learning with noisy labels as
  semi-supervised learning. arXiv preprint arXiv:200207394

\bibitem[{Li et~al(2021)Li, Liu, Han, Niu, and Sugiyama}]{li2021provably}
Li X, Liu T, Han B, Niu G, Sugiyama M (2021) Provably end-to-end label-noise
  learning without anchor points. In: ICML

\bibitem[{Liu et~al(2020)Liu, Niles-Weed, Razavian, and
  Fernandez-Granda}]{liu2020early}
Liu S, Niles-Weed J, Razavian N, Fernandez-Granda C (2020) Early-learning
  regularization prevents memorization of noisy labels. NeurIPS

\bibitem[{Liu and Guo(2020)}]{liu2020peer}
Liu Y, Guo H (2020) Peer loss functions: Learning from noisy labels without
  knowing noise rates. In: ICML

\bibitem[{Loshchilov and Hutter(2018)}]{loshchilov2018decoupled}
Loshchilov I, Hutter F (2018) Decoupled weight decay regularization. In: ICLR

\bibitem[{Lukasik et~al(2020)Lukasik, Bhojanapalli, Menon, and
  Kumar}]{lukasik2020does}
Lukasik M, Bhojanapalli S, Menon A, Kumar S (2020) Does label smoothing
  mitigate label noise? In: ICML

\bibitem[{Ma et~al(2018)Ma, Wang, Houle, Zhou, Erfani, Xia, Wijewickrema, and
  Bailey}]{ma2018dimensionality}
Ma X, Wang Y, Houle ME, Zhou S, Erfani S, Xia S, Wijewickrema S, Bailey J
  (2018) Dimensionality-driven learning with noisy labels. In: ICML

\bibitem[{Miyato et~al(2018)Miyato, Maeda, Koyama, and
  Ishii}]{miyato2018virtual}
Miyato T, Maeda Si, Koyama M, Ishii S (2018) Virtual adversarial training: a
  regularization method for supervised and semi-supervised learning. TPAMI

\bibitem[{Natarajan et~al(2013)Natarajan, Dhillon, Ravikumar, and
  Tewari}]{natarajan2013learning}
Natarajan N, Dhillon IS, Ravikumar PK, Tewari A (2013) Learning with noisy
  labels. NeurIPS 26

\bibitem[{Nishi et~al(2021)Nishi, Ding, Rich, and
  Hollerer}]{nishi2021augmentation}
Nishi K, Ding Y, Rich A, Hollerer T (2021) Augmentation strategies for learning
  with noisy labels. In: CVPR

\bibitem[{Paszke et~al(2019)Paszke, Gross, Massa, Lerer, Bradbury, Chanan,
  Killeen, Lin, Gimelshein, Antiga et~al}]{paszke2019pytorch}
Paszke A, Gross S, Massa F, Lerer A, Bradbury J, Chanan G, Killeen T, Lin Z,
  Gimelshein N, Antiga L, et~al (2019) Pytorch: An imperative style,
  high-performance deep learning library. NeurIPS

\bibitem[{Patrini et~al(2017)Patrini, Rozza, Krishna~Menon, Nock, and
  Qu}]{patrini2017making}
Patrini G, Rozza A, Krishna~Menon A, Nock R, Qu L (2017) Making deep neural
  networks robust to label noise: A loss correction approach. In: CVPR

\bibitem[{Rasmus et~al(2015)Rasmus, Berglund, Honkala, Valpola, and
  Raiko}]{rasmus2015semi}
Rasmus A, Berglund M, Honkala M, Valpola H, Raiko T (2015) Semi-supervised
  learning with ladder networks. NeurIPS

\bibitem[{Reed et~al(2014)Reed, Lee, Anguelov, Szegedy, Erhan, and
  Rabinovich}]{reed2014training}
Reed S, Lee H, Anguelov D, Szegedy C, Erhan D, Rabinovich A (2014) Training
  deep neural networks on noisy labels with bootstrapping. arXiv preprint
  arXiv:14126596

\bibitem[{Sajjadi et~al(2016)Sajjadi, Javanmardi, and
  Tasdizen}]{sajjadi2016regularization}
Sajjadi M, Javanmardi M, Tasdizen T (2016) Regularization with stochastic
  transformations and perturbations for deep semi-supervised learning. NeurIPS

\bibitem[{Shalev-Shwartz and Ben-David(2014)}]{shalev2014understanding}
Shalev-Shwartz S, Ben-David S (2014) Understanding machine learning: From
  theory to algorithms. Cambridge university press

\bibitem[{Shen and Sanghavi(2019)}]{shen2019learning}
Shen Y, Sanghavi S (2019) Learning with bad training data via iterative trimmed
  loss minimization. In: ICML

\bibitem[{Sohn et~al(2020)Sohn, Berthelot, Carlini, Zhang, Zhang, Raffel,
  Cubuk, Kurakin, and Li}]{sohn2020fixmatch}
Sohn K, Berthelot D, Carlini N, Zhang Z, Zhang H, Raffel CA, Cubuk ED, Kurakin
  A, Li CL (2020) Fixmatch: Simplifying semi-supervised learning with
  consistency and confidence. NeurIPS

\bibitem[{Sun et~al(2022)Sun, Shen, Huang, Wang, Shu, Yao, and
  Tang}]{sun2022pnp}
Sun Z, Shen F, Huang D, Wang Q, Shu X, Yao Y, Tang J (2022) Pnp: Robust
  learning from noisy labels by probabilistic noise prediction. In: CVPR

\bibitem[{Tanaka et~al(2018)Tanaka, Ikami, Yamasaki, and
  Aizawa}]{tanaka2018joint}
Tanaka D, Ikami D, Yamasaki T, Aizawa K (2018) Joint optimization framework for
  learning with noisy labels. In: CVPR

\bibitem[{Tarvainen and Valpola(2017)}]{tarvainen2017mean}
Tarvainen A, Valpola H (2017) Mean teachers are better role models:
  Weight-averaged consistency targets improve semi-supervised deep learning
  results. NeurIPS

\bibitem[{Wang et~al(2022)Wang, Chen, Heng, Hou, Savvides, Shinozaki, Raj, Wu,
  and Wang}]{wang2022freematch}
Wang Y, Chen H, Heng Q, Hou W, Savvides M, Shinozaki T, Raj B, Wu Z, Wang J
  (2022) Freematch: Self-adaptive thresholding for semi-supervised learning.
  arXiv preprint arXiv:220507246

\bibitem[{Wei et~al(2020)Wei, Feng, Chen, and An}]{wei2020combating}
Wei H, Feng L, Chen X, An B (2020) Combating noisy labels by agreement: A joint
  training method with co-regularization. In: CVPR

\bibitem[{Wei and Liu(2020)}]{wei2020optimizing}
Wei J, Liu Y (2020) When optimizing $ f $-divergence is robust with label
  noise. arXiv preprint arXiv:201103687

\bibitem[{Wei et~al(2021{\natexlab{a}})Wei, Liu, Liu, Niu, and
  Liu}]{wei2021understanding}
Wei J, Liu H, Liu T, Niu G, Liu Y (2021{\natexlab{a}}) Understanding
  (generalized) label smoothing whenlearning with noisy labels. arXiv preprint
  arXiv:210604149

\bibitem[{Wei et~al(2021{\natexlab{b}})Wei, Zhu, Cheng, Liu, Niu, and
  Liu}]{wei2021learning}
Wei J, Zhu Z, Cheng H, Liu T, Niu G, Liu Y (2021{\natexlab{b}}) Learning with
  noisy labels revisited: A study using real-world human annotations. arXiv
  preprint arXiv:211012088

\bibitem[{Wolpert and Macready(1997)}]{wolpert1997no}
Wolpert DH, Macready WG (1997) No free lunch theorems for optimization. TEVC

\bibitem[{Xia et~al(2019)Xia, Liu, Wang, Han, Gong, Niu, and
  Sugiyama}]{xia2019anchor}
Xia X, Liu T, Wang N, Han B, Gong C, Niu G, Sugiyama M (2019) Are anchor points
  really indispensable in label-noise learning? NeurIPS

\bibitem[{Xu et~al(2019)Xu, Cao, Kong, and Wang}]{xu2019l_dmi}
Xu Y, Cao P, Kong Y, Wang Y (2019) L\_dmi: A novel information-theoretic loss
  function for training deep nets robust to label noise. NeurIPS

\bibitem[{Xu et~al(2021)Xu, Shang, Ye, Qian, Li, Sun, Li, and Jin}]{xu2021dash}
Xu Y, Shang L, Ye J, Qian Q, Li YF, Sun B, Li H, Jin R (2021) Dash:
  Semi-supervised learning with dynamic thresholding. In: ICML, pp 11525--11536

\bibitem[{Yang et~al(2021)Yang, Song, King, and Xu}]{yang2021survey}
Yang X, Song Z, King I, Xu Z (2021) A survey on deep semi-supervised learning.
  arXiv preprint arXiv:210300550

\bibitem[{Yao et~al(2021)Yao, Sun, Zhang, Shen, Wu, Zhang, and
  Tang}]{yao2021jo}
Yao Y, Sun Z, Zhang C, Shen F, Wu Q, Zhang J, Tang Z (2021) Jo-src: A
  contrastive approach for combating noisy labels. In: CVPR

\bibitem[{Yi and Wu(2019)}]{yi2019probabilistic}
Yi K, Wu J (2019) Probabilistic end-to-end noise correction for learning with
  noisy labels. In: CVPR

\bibitem[{Yu et~al(2019)Yu, Han, Yao, Niu, Tsang, and Sugiyama}]{yu2019does}
Yu X, Han B, Yao J, Niu G, Tsang I, Sugiyama M (2019) How does disagreement
  help generalization against label corruption? In: ICML

\bibitem[{Zhang et~al(2021)Zhang, Wang, Hou, Wu, Wang, Okumura, and
  Shinozaki}]{zhang2021flexmatch}
Zhang B, Wang Y, Hou W, Wu H, Wang J, Okumura M, Shinozaki T (2021) Flexmatch:
  Boosting semi-supervised learning with curriculum pseudo labeling. NeurIPS 34

\bibitem[{Zhang et~al(2017)Zhang, Cisse, Dauphin, and
  Lopez-Paz}]{zhang2017mixup}
Zhang H, Cisse M, Dauphin YN, Lopez-Paz D (2017) mixup: Beyond empirical risk
  minimization. arXiv preprint arXiv:171009412

\bibitem[{Zhang and Sabuncu(2018)}]{zhang2018generalized}
Zhang Z, Sabuncu M (2018) Generalized cross entropy loss for training deep
  neural networks with noisy labels. NeurIPS

\bibitem[{Zhao et~al(2022)Zhao, Li, Qin, Liu, and Yu}]{zhao2022centrality}
Zhao G, Li G, Qin Y, Liu F, Yu Y (2022) Centrality and consistency: two-stage
  clean samples identification for learning with instance-dependent noisy
  labels. arXiv preprint arXiv:220714476

\bibitem[{Zhu et~al(2021)Zhu, Liu, and Liu}]{zhu2021second}
Zhu Z, Liu T, Liu Y (2021) A second-order approach to learning with
  instance-dependent label noise. In: CVPR

\end{thebibliography}


\end{document}